\documentclass[letterpaper,journal]{IEEEtran}

\usepackage{amsmath,amsfonts,amssymb}
\usepackage{array}
\usepackage{textcomp}
\usepackage{stfloats}
\usepackage[caption=false,font=footnotesize]{subfig}
\usepackage{url}
\usepackage{verbatim}
\usepackage{graphicx}
\usepackage{balance}
\usepackage{multirow}
\usepackage{booktabs}
\usepackage{xcolor}
\usepackage{enumitem}
\usepackage{algorithm}
\usepackage{algorithmic}

\def\BibTeX{{\rm B\kern-.05em{\sc i\kern-.025em b}\kern-.08em
T\kern-.1667em\lower.7ex\hbox{E}\kern-.125emX}}

\begin{document}

\title{Toward Universal Skeleton-Based Action Recognition across Heterogeneous Skeletons and Open Vocabularies}

\author{Jidong Kuang, Hongsong Wang,
Jie Gui,~\IEEEmembership{Senior Member,~IEEE},
Yuan Yan Tang,~\IEEEmembership{Life Fellow,~IEEE},
and James Tin-Yau Kwok,~\IEEEmembership{Fellow,~IEEE}
\thanks{This work was supported in part by the grant of the CIPS-SMP-Zhipu Large Model Fund. We thank the Big Data Computing Center of Southeast University for providing the facility support on the numerical calculations. (Corresponding authors: Jie Gui and Hongsong Wang.)}
\thanks{Jidong Kuang is with the School of Cyber Science and Engineering, Southeast University, Nanjing 211189, China (e-mail: jidongkuang@seu.edu.cn).}
\thanks{Hongsong Wang is with the School of Computer Science and Engineering, Southeast University, Nanjing 211102, China (e-mail: hongsongwang@seu.edu.cn).}
\thanks{Jie Gui is with the School of Cyber Science and Engineering, Southeast University, Nanjing 211189, China, also with Purple Mountain Laboratories, Nanjing 211111, China, and also with the Engineering Research Center of Blockchain Application, Supervision and Management (Southeast University), Ministry of Education, Nanjing 210000, China (e-mail: guijie@seu.edu.cn).}
\thanks{Yuan Yan Tang is with the Department of Computer and Information Science, University of Macau, Macau, China, and also with the Faculty of Science and Technology, UOW College Hong Kong, Hong Kong, China (e-mail: yytang@um.edu.mo).}
\thanks{James Tin-Yau Kwok is with the Department of Computer Science and Engineering, The Hong Kong University of Science and Technology, Hong Kong, China (e-mail: jamesk@cse.ust.hk).}}

\markboth{IEEE Transactions on Image Processing}%
{Kuang \MakeLowercase{\textit{et al.}}: Toward Universal Skeleton-Based Action Recognition across Heterogeneous Skeletons and Open Vocabularies}

\maketitle

\begin{abstract}
Skeleton data used for action recognition are acquired from a wide range of sources, including depth sensors, marker-based motion capture systems, and 2D/3D pose estimators. These sources yield skeletons that differ in joint number, skeletal topology, and coordinate dimensionality, making skeleton data inherently heterogeneous. However, previous works overlook the data heterogeneity of skeletons and solely construct models using homogeneous skeletons. Moreover, open-vocabulary action recognition is also essential for real-world applications. To this end, this work studies the challenging problem of heterogeneous skeleton-based action recognition with open vocabularies. We construct a large-scale Heterogeneous Open-Vocabulary (HOV) Skeleton dataset by integrating and refining multiple representative large-scale skeleton-based action datasets. To address universal skeleton-based action recognition, we propose a Transformer-based model that standardizes heterogeneous skeletons into a unified representation, encodes multi-modal skeleton embeddings with a two-stream motion encoder to learn spatio-temporal action representations, and maps them to a semantic space through multi-grained motion-text alignment. The alignment incorporates contrastive learning at three levels: global instance alignment, stream-specific alignment, and fine-grained alignment. Extensive experiments on popular benchmarks with heterogeneous skeleton data demonstrate both the effectiveness and the generalization ability of the proposed method. Code is available at \url{https://github.com/jidongkuang/Universal-Skeleton}.
\end{abstract}

\begin{IEEEkeywords}
Skeleton-based action recognition, heterogeneous skeletons.
\end{IEEEkeywords}

\section{Introduction}
\label{sec:introduction}

\IEEEPARstart{A}{ction} recognition is a fundamental task in computer vision and has become increasingly crucial with the development of intelligent systems. Action recognition enhances human-robot interaction, allowing robots to understand human gestures, postures, and intentions. It also boosts the agents' adaptability to diverse scenarios, and promotes their intelligence by enabling them to analyze human behaviors and make autonomous decisions. Action recognition also holds a significant role in other fields such as medical rehabilitation, healthcare, smart homes, intelligent education and intelligent sports.

\begin{figure}[t]
    \centering
    \includegraphics[width=\linewidth]{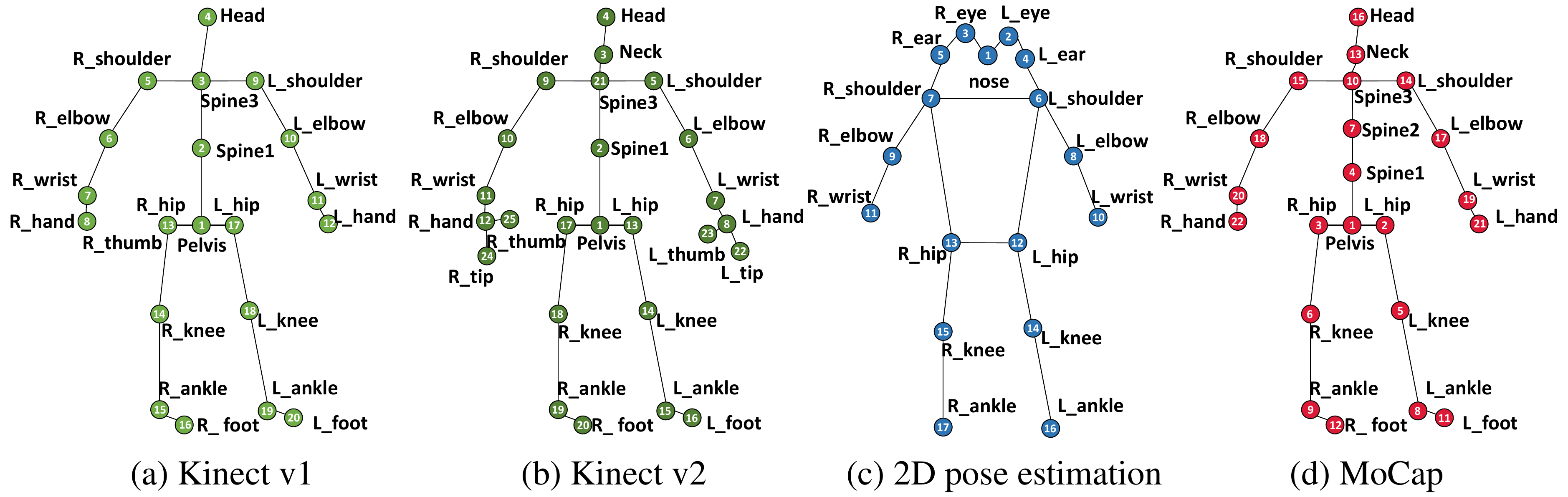}
    \caption{Comparison of heterogeneous skeletons from various common sources, illustrating differences in joint count and structure. (a) Kinect v1 (20 joints), (b) Kinect v2 (25 joints), (c) 2D pose estimation output (e.g., 17 joints), (d) MoCap data adhering to a model like SMPL (e.g., 22 joints).}
    \label{fig:heterogeneous}
    \vspace{-3mm}
\end{figure}

Skeletons are a prevalent modality in action recognition tasks. In contrast to alternative modalities like video sequences and depth maps, skeletons focus on essential joint positions, offering high abstraction. This results in low-complexity data, reducing computational burden. Another key advantage is their robustness. Unlike video and depth map data, which can be degraded by lighting changes, occlusions, or background clutter, skeleton data remains reliable. Moreover, the inherent alignment of skeleton data with human kinematics facilitates a more precise representation of motion. These characteristics make skeletons an ideal choice for action recognition.

Although deep learning methods have advanced rapidly in skeleton-based human action recognition, these methods primarily focus on proposing different network architectures for single homogeneous data. Based on different types of neural networks, they can be broadly classified into three categories: Recurrent Neural Networks (RNNs)-based methods~\cite{du2016representation,wang2018beyond}, Graph Convolutional Networks (GCNs)-based methods~\cite{yan2018spatial}, and Transformers-based methods~\cite{zhu2023motionbert}. These methods overlook the data heterogeneity of skeletons, which arises from a multitude of motion capture technologies, mountable sensors, and pose estimation algorithms used in data collection. These diverse sources yield skeleton data with varying numbers of joints, disparate skeletal topologies, and different coordinate dimensionalities. As shown in Figure~\ref{fig:heterogeneous}, the Kinect v2 depth sensor captures 25 3D joints, while 2D pose estimation algorithms applied to RGB videos typically yield 17 2D joints. Additionally, motion capture (MoCap) systems following biomechanical models such as SMPL~\cite{loper2015smpl} produce skeletons with 22 3D joints.

Moreover, most existing action recognition paradigms are designed under the closed-set assumption. Although self-supervised skeleton-based action recognition~\cite{guo2022contrastive,sun2023unified} and zero-shot skeleton-based action recognition~\cite{zhou2023zero,zhu2024part} can be applied in open-set scenarios, the former requires an additional fine-tuning step, while the latter is restricted to a limited number of unseen classes. Large-scale skeleton-based action recognition with open vocabularies that go beyond explicitly trained categories is an important problem that needs to be solved. However, in realistic applications, data heterogeneity further complicates the goal of achieving all-in-one skeleton-based action recognition. In such a case, the open-vocabulary system must generalize not only to novel semantic concepts but also to the diverse and heterogeneous skeletal data associated with them.

To address this unified challenge, this work studies heterogeneous skeleton-based action recognition with open vocabularies. The objective is to design a unified action recognition model capable of coping with heterogeneous skeletons under an open-set setting. To address this task, we introduce the Heterogeneous Open-Vocabulary (HOV) Skeleton dataset, which integrates several representative large-scale heterogeneous skeleton-based action datasets. Among them, we establish a new multi-label classification benchmark on HumanML3D to advance action recognition on MoCap data. We propose a Transformer-based architecture that enables unified representation learning across diverse heterogeneous skeleton data through multi-grained motion-text alignment. This alignment strategy based on contrastive learning incorporates three levels: global instance alignment, stream-specific alignment (SSA), and fine-grained alignment (FGA). By capturing information at multiple granularities, our approach enhances the recognition of rare actions, thereby promoting robust heterogeneous skeleton recognition in open-vocabulary scenarios.

Our contributions can be summarized as follows:
\begin{itemize}
\item We study the problem of universal skeleton-based action recognition, which jointly addresses the critical challenges of data heterogeneity and open vocabularies.
\item We introduce the large-scale HOV Skeleton dataset to accommodate the task of heterogeneous skeleton-based action recognition with open vocabularies.
\item We propose a Transformer-based architecture with multi-grained motion-text alignment for all-in-one skeleton-based action recognition.
\end{itemize}

\section{Related Work}

\subsection{3D Action Representation Learning}
3D action representation learning primarily relies on the self-supervised paradigm and is applied to various downstream tasks. In early studies~\cite{xu2021unsupervised,wang2022contrast}, researchers commonly employ encoder-decoder architectures to design self-supervised learning objectives. Contrastive learning currently dominates as the mainstream paradigm for self-supervised action representation learning~\cite{wu2024scd,lin2023actionlet}. Lin et al.~\cite{lin2024idempotent} enforce feature space consistency through an idempotent generative model (IGM). Emerging techniques further push the boundaries of performance. Gong et al.~\cite{gong2025rethinking} demonstrate the superiority of hybrid strategies by integrating masked joint reconstruction with contrastive objectives. However, existing skeleton representation methods are constrained by specific data topologies, lacking a topology-independent mechanism for generalizable representation.

\subsection{Zero-Shot Skeleton-Based Action Recognition}
Zero-shot skeleton-based action recognition (ZSAR) classifies unseen action categories by establishing visual-semantic space alignment~\cite{chen2025neuron,do2025bridging,wu2025frequency}. Early approaches employ basic semantic constraints to build cross-modal associations. Gupta et al.~\cite{gupta2021syntactically} align skeleton sequences with linguistic descriptions through Parts of Speech (PoS) tagging mechanisms. Zhou et al.~\cite{zhou2023zero} establish global visual-semantic distribution alignment by introducing mutual information estimation and maximization. The advent of Large Language Models (LLMs) drives semantic enhancement strategies to the forefront. Zhu et al.~\cite{zhu2024part} leverage LLMs to generate fine-grained motion descriptions, achieving local and global alignment through adaptive joint feature grouping. Sinha et al.~\cite{sinha2025ski} distill skeleton information into Vision Language Models (VLMs) and perform zero-shot recognition using the RGB modality during inference. However, most of these methods rely on pre-trained skeleton encoders~\cite{yan2018spatial}, resulting in a disconnect between feature extraction and cross-modal alignment. This limitation restricts cross-dataset transfer capabilities, particularly when handling heterogeneous skeletons.

\subsection{Unified Skeleton-Based Action Recognition}
Unified skeleton-based action recognition aims to employ a single model to handle diverse action recognition scenarios. Only a few works have addressed this challenging problem, which can be categorized into three types: skeleton unification, domain unification, and semantic unification. We briefly review works of the first two types, where semantic unification corresponds to open-vocabulary or zero-shot settings, reviewed above. For skeleton unification, UPS~\cite{foo2023unified} represents both pose coordinates and action labels as language sequences. Wang et al.~\cite{wang2025heterogeneous} explore heterogeneous skeleton data to enhance action representations. For domain unification, LLM-AR~\cite{qu2024llms} projects skeleton sequences into tokens and leverages large language models for recognition. Furthermore, UNIK~\cite{yang2021unik} achieves cross-domain generalization by learning dependency matrices via multi-head attention. To address multi-participant scenarios, SkeleTR~\cite{duan2023skeletr} adopts a two-stage paradigm that integrates graph convolutional networks with Transformers. Despite these advances toward unified modeling, existing frameworks are typically restricted to a single skeleton format or a fixed label space, and none addresses both simultaneously.

\section{HOV Skeleton Dataset}

\begin{table}[t]
\centering
\caption{Comparison of problem settings. ``Formats'': the number of skeleton
formats a single trained model accepts. ``Models'': the number of separate
models required to cover $n$ evaluation datasets. ``Cls. head'': whether
recognition relies on a dataset-specific classification layer. ``Open
vocab.'': whether novel categories can be recognized without modifying the
model. Among the four formats our model accepts, Kinect v1 is never observed
during training.}
\label{tab:setting_comparison}
\resizebox{\columnwidth}{!}{
\begin{tabular}{lcccc}
\toprule
Paradigm & Formats & Models & Cls. head & Open vocab. \\
\midrule
Supervised~\cite{lee2023hierarchically,zhou2024blockgcn,liu2025revealing} & 1 & $n$ & \checkmark & $\times$ \\
Self-supervised~\cite{guo2022contrastive,sun2023unified,gong2025rethinking} & 1 & $n$ & \checkmark & $\times$ \\
Zero-shot~\cite{zhou2023zero,zhu2024part,chen2025neuron} & 1 & $n$ & $\times$ & partial \\
Unified modeling~\cite{foo2023unified,yang2021unik,qu2024llms,wang2025heterogeneous} & 1--2 & 1 & \checkmark & $\times$ \\
\midrule
Ours & \textbf{4} & \textbf{1} & $\times$ & \checkmark \\
\bottomrule
\end{tabular}}
\vspace{-3mm}
\end{table}

Existing skeleton-based action recognition methods assume a closed set of
categories and a single predefined skeleton format, and therefore struggle
with diverse actions across varying scenarios.
Table~\ref{tab:setting_comparison} summarizes this landscape. Supervised
and self-supervised methods attach a dataset-specific classification layer
to each evaluation dataset. Zero-shot methods remove the classification
layer but remain tied to one skeleton format and are typically evaluated
on a few held-out classes within a single dataset. Unified modeling
approaches relax one constraint while retaining the other, sharing one
backbone across tasks or across a small number of formats yet still
predicting over a label space fixed at training time. In contrast, the
setting studied here relaxes both constraints at once: a single set of
weights accepts four skeleton formats, and recognition is performed by
matching visual features against text embeddings, so neither the input
format nor the label space is fixed in advance. To support this setting,
we introduce the Heterogeneous Open-Vocabulary (HOV) Skeleton dataset,
which integrates and refines multiple representative large-scale
skeleton-based action datasets and offers two distinctive
characteristics: heterogeneous skeletons and open vocabularies.

\subsection{Heterogeneous Skeletons}
Human skeleton data acquired from different sources exhibit inherent structural heterogeneity. This variability manifests in the number of joints, topology, and dimensionality. For example, early depth sensors like Kinect v1 typically provide 20 3D joints, while its successor Kinect v2 captures 25 3D joints. In contrast, 2D pose estimation from RGB videos commonly yields 17 2D joints~\cite{duan2022revisiting}. Motion capture (MoCap) systems may follow biomechanical models such as SMPL~\cite{loper2015smpl}, which comprise around 22 3D joints. This inherent heterogeneity highlights the need for a unified input processing strategy to facilitate cross-dataset learning.

\begin{figure}[t]
    \centering
    \begin{minipage}[t]{0.36\linewidth}
        \centering
        \includegraphics[width=0.9\linewidth,height=0.9\linewidth]{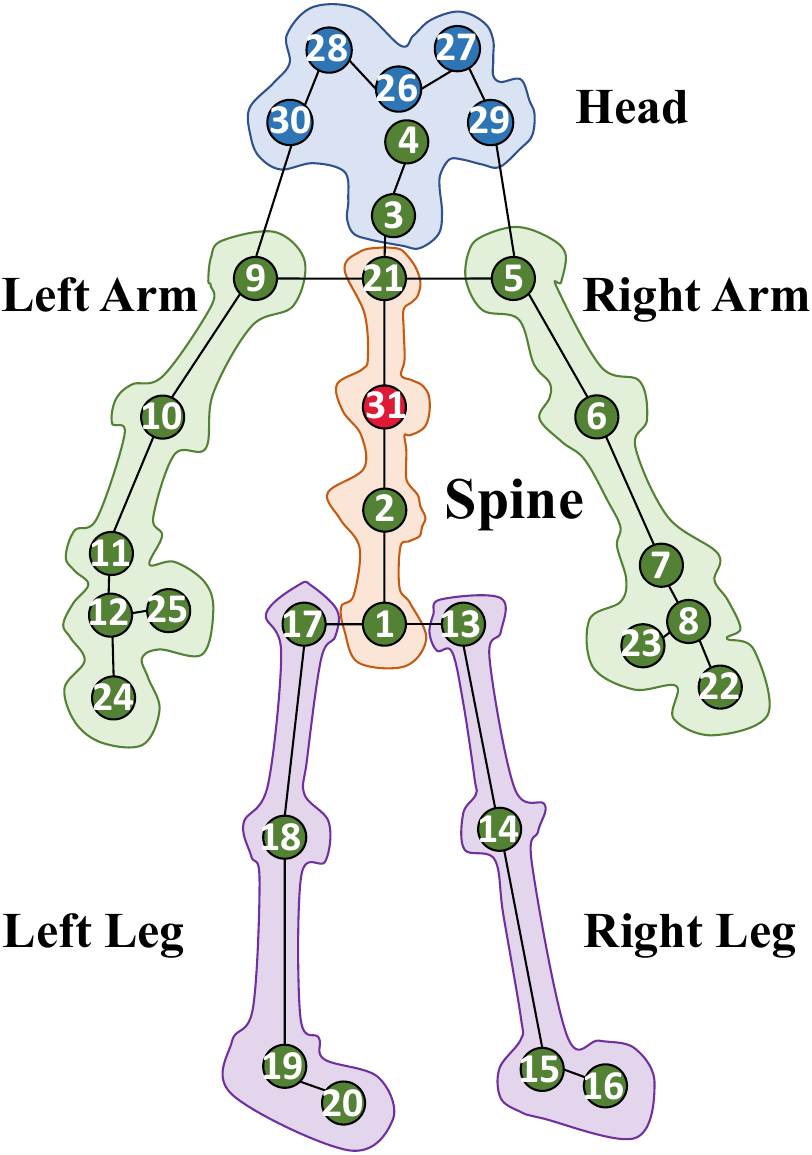}
        (a)
    \end{minipage}
    \begin{minipage}[t]{0.6\linewidth}
        \centering
        \includegraphics[width=\linewidth]{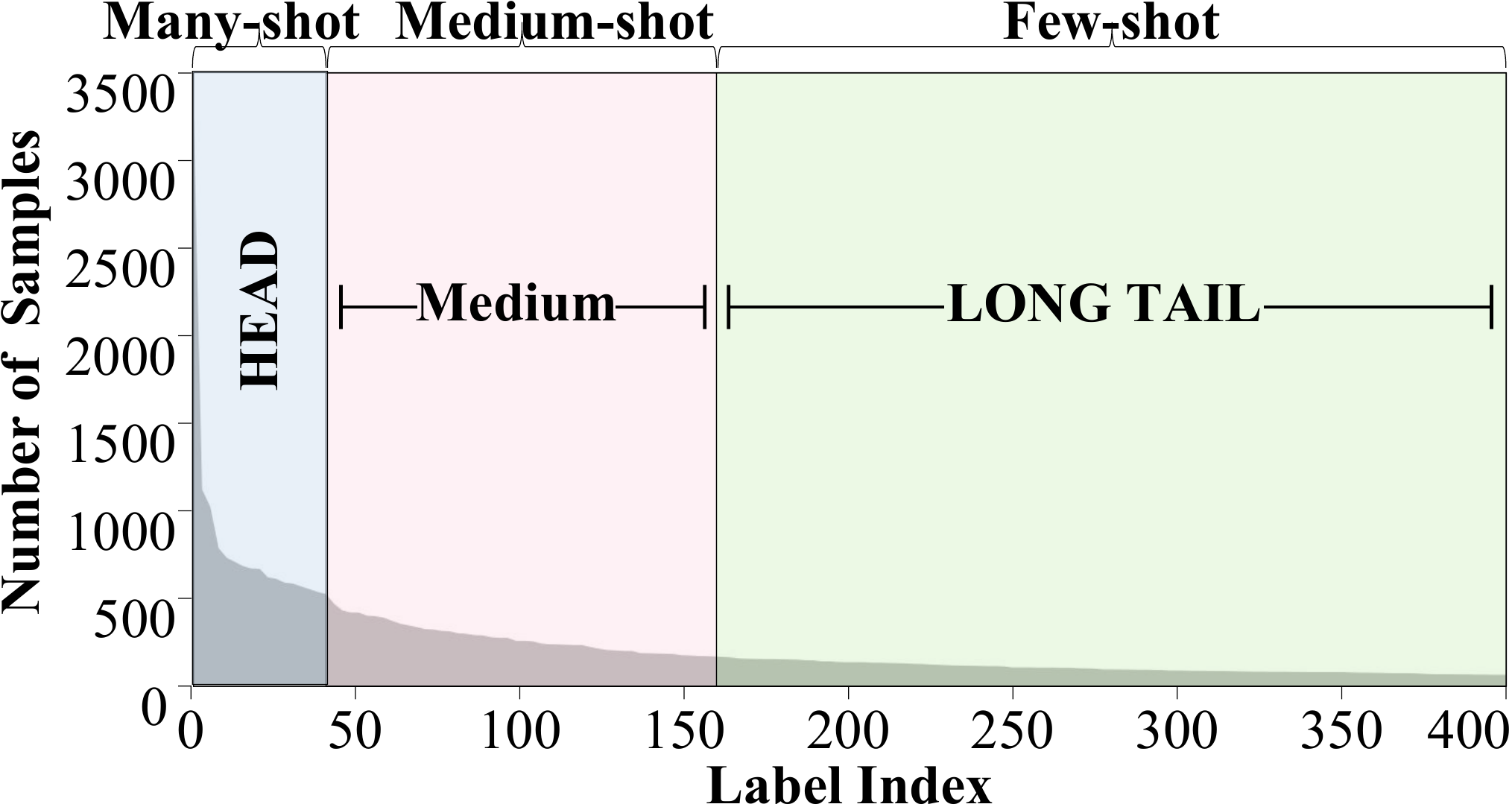}
        (b)
    \end{minipage}
    \caption{Characteristics of skeleton structure and sample distribution of the HOV Skeleton dataset: (a) Definition of the human skeleton with joint numbering for the HOV Skeleton dataset. (b) Long-tailed distribution of action categories on HumanML3D. The three tasks are many-shot, medium-shot, and few-shot action recognition, respectively.}
    \label{fig:unified_skeleton}
    \vspace{-3mm}
\end{figure}

To address the problem of heterogeneous skeletons, the HOV Skeleton integrates several representative heterogeneous skeleton datasets, including NW-UCLA~\cite{wang2014cross}, 3D and 2D versions of NTU-60~\cite{shahroudy2016ntu} and NTU-120~\cite{liu2019ntu}, and HumanML3D~\cite{guo2022generating}. For the HOV Skeleton dataset, human joint numbers are shown in Figure~\ref{fig:unified_skeleton}(a). However, as HumanML3D is a large-scale dataset primarily designed for motion synthesis, its original data and annotations are not directly applicable to action recognition. Therefore, we conduct relevant manual processing for this dataset, and the detailed process is presented in the subsequent descriptions.

\subsection{Benchmark Construction}
\label{sec:dataset2}
Open-vocabulary action recognition enables models to recognize novel action categories. Advancing this field requires both large-scale datasets covering diverse actions and realistic complexity, including data heterogeneity. However, there is a lack of a suitable large-scale benchmark for open-vocabulary skeleton-based action recognition.

Since the original annotations contain multiple sub-actions within a single motion sequence, in order to adapt them to the task of action recognition, we manually process these annotations and then establish a novel multi-label classification benchmark on HumanML3D. By leveraging the rich motion data of HumanML3D, we aim to facilitate action recognition on the MoCap data. The construction process of the benchmark involves several key steps:

\noindent\textbf{Description-to-Label Conversion.}
We leverage LLMs to extract core action verbs from text descriptions following~\cite{bensabath2024cross}. For example, ``a man squats extraordinarily low then bolts up in an unsatisfactory jump'' is converted to ``squat and jump,'' and then tokenized into a label list: [`squat', `jump']. Each description undergoes this process, resulting in 3 to 4 labels per sequence on average.

\noindent\textbf{Label Space Clustering.}
We collect all unique action tokens across the dataset, yielding an initial label space of 8,801 classes. However, this label space suffers from semantic redundancy (e.g., ``walk'', ``walks'', ``walking'') and an extreme long-tailed distribution. To mitigate these issues, we apply the Balanced K-Means algorithm to cluster the original labels into 400 semantically coherent categories. This step merges rare and semantically similar categories into a more compact and balanced label set. The final multi-label annotation for each sequence is defined as the union of its refined label candidates.

\noindent\textbf{Stratified Dataset Partition.}
The original split (80\% training, 15\% test, 5\% validation) is designed for motion synthesis. We reorganize the dataset into 70\% training and 30\% testing to align with classification benchmarks, using stratified sampling to maintain category distribution consistency.

\noindent\textbf{Long-Tailed Evaluation.}
Frequency analysis reveals a long-tailed distribution among the 400 categories, as shown in Figure~\ref{fig:unified_skeleton}(b). Specifically, a small number of head classes contain a large volume of samples, while the majority of tail classes have only a few. To systematically assess model performance across classes of varying frequencies, we categorize the labels into three subsets: head (top 10\%), medium (middle 30\%), and tail (bottom 60\%). During evaluation, we report performance on the overall set and each subset to show the model's capability in handling rare categories.

Based on this process, we establish a new multi-label action classification benchmark on HumanML3D, featuring 400 categories with a long-tailed distribution. Its large and fine-grained label space further supports open-vocabulary evaluation, where novel categories are held out at training time and recognized through motion-text matching. Further details on our dataset's construction, including its comparison with related benchmark BABEL~\cite{punnakkal2021babel}, are provided in the supplementary material.

\begin{figure*}[t]
    \centering
    \includegraphics[width=0.9\linewidth]{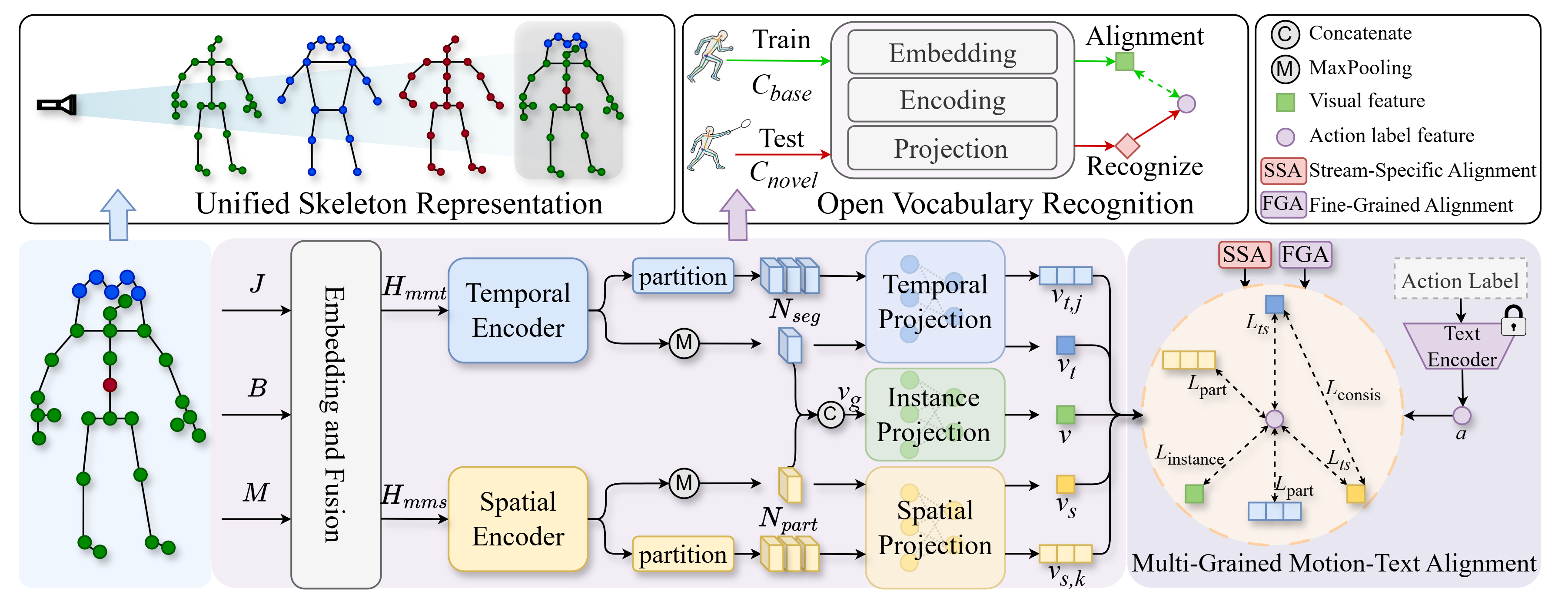}
    \caption{Overview of the proposed architecture for heterogeneous skeleton-based action recognition with open vocabularies. Heterogeneous skeletons from diverse sources are standardized into a unified representation and subsequently processed by parallel Transformer-based temporal and spatial encoders. The visual features after decoupled projection are subjected to a multi-grained alignment strategy to facilitate open-vocabulary recognition. The Stream-Specific Alignment (SSA) module captures action information from distinct streams while maintaining spatio-temporal consistency. The Fine-Grained Alignment (FGA) module enhances the understanding of action composition, thereby improving the recognition capabilities for rare actions. Through the multi-grained alignment strategy, visual features are aligned with action label embeddings from a text encoder, enabling open-vocabulary capabilities.}
    \label{fig:method}
    \vspace{-3mm}
\end{figure*}

\section{Universal Model for Action Recognition}
\label{sec:method}

To address the issues of both data heterogeneity and open vocabularies, we design a network for universal skeleton-based action recognition. The overall architecture of the proposed model is illustrated in Figure~\ref{fig:method}. Our approach focuses on unifying heterogeneous skeletons and employing a multi-grained cross-modal alignment mechanism to enable open-vocabulary recognition.

\subsection{Unified Skeleton Representation}
\label{sec:Unified}

To handle heterogeneous skeletons, a crucial first step is to standardize these disparate inputs into a unified format. Let an input skeleton sequence be represented as $\mathbf{X}_{\text{orig}} \in \mathbb{R}^{N \times C \times T \times K \times M}$, where $N$ is the batch size, and $C, T, K, M$ denote the number of coordinate dimensions, frames, joints, and members, respectively. We define a unified spatial structure by establishing maximum joint and member counts, denoted as $K_{\text{unified}}$ and $M_{\text{unified}}$. Let $\mathcal{D}$ denote the set of supported source formats.
$K_{\text{unified}}$ is the number of distinct joints across all formats,
and $M_{\text{unified}}$ is the maximum number of members among them:
\begin{equation}
    K_{\text{unified}} = \Bigl|\scalebox{0.8}{$\bigcup$}_{d \in \mathcal{D}}
    \mathcal{K}_d\Bigr|, \quad
    M_{\text{unified}} = \max_{d \in \mathcal{D}} M_d,
    \label{eq:unified_k}
\end{equation}
where $\mathcal{K}_d$ denotes the set of distinct joints in format $d$, and $M_d$ denotes the maximum number of individuals per sequence in format $d$. For an input sequence whose joint count is smaller than $K_{\text{unified}}$, we expand it to the unified spatial structure through kinematic 
imputation. For a missing joint with a reliable anatomical correspondence to observed joints, we impute it using human-body priors: trunk joints are interpolated as midpoints of observed joint pairs, and terminal joints are extrapolated along their kinematic chains with fixed ratios. The member dimension is expanded in a similar manner. This expansion ensures that all skeletons conform to a consistent spatial structure of $K_{\text{unified}} \times M_{\text{unified}}$.

\subsection{Motion Encoder for Skeletons}
The proposed model architecture comprises three core stages: multi-modal input embedding and fusion, spatio-temporal feature extraction, and projection layers for cross-modal alignment.

\noindent\textbf{Multi-Modal Embedding and Fusion.}
To capture diverse aspects of human motion, we derive multiple modalities from the unified joint coordinates: joint $\mathcal{J}$, bone $\mathcal{B}$, and motion $\mathcal{M}$. The bone $\mathcal{B}$ explicitly models the body's kinematic structure by computing the differences between connected joints, while the motion $\mathcal{M}$ captures temporal dynamics by calculating frame-to-frame joint displacements. Each derived modality undergoes modality-specific processing before fusion. We employ distinct embedding modules implemented as MLPs to map the flattened representations of each modality into a common hidden space with dimension $D_h$. These representations are prepared for both temporal and spatial processing streams. The embedded features $\mathbf{H}_{\mathcal{J}}, \mathbf{H}_{\mathcal{B}}, \mathbf{H}_{\mathcal{M}}$ are fused via weighted averaging followed by linear projection, yielding the fused multi-modal representations for the temporal stream $\mathbf{H}_{mmt} \in \mathbb{R}^{N \times T \times D_h}$ and the spatial stream $\mathbf{H}_{mms} \in \mathbb{R}^{N \times (K_{\text{unified}} \times M_{\text{unified}}) \times D_h}$.

\noindent\textbf{Spatio-Temporal Encoding.}
A two-stream Transformer-based backbone processes the fused features through parallel temporal and spatial encoders. The Temporal Encoder $\mathcal{E}_{t}$ operates on the fused temporal representation $\mathbf{H}_{mmt}$. Standard sinusoidal positional encodings $\mathbf{E}_{pos}$ are added to infuse temporal order information. The sequence is then passed through a stack of $L$ Transformer encoder layers, designed to capture the temporal evolution and dynamics of the action. In parallel, the Spatial Encoder $\mathcal{E}_{s}$ processes the fused spatial representation $\mathbf{H}_{mms}$. Learnable spatial embeddings $\mathbf{E}_{spe}$ are added to the input to encode spatial position information. The sequence subsequently passes through $L$ Transformer encoder layers, modeling the complex inter-relationships and configurations between joints across the unified skeleton structure. The backbone yields sequence-level feature representations for both streams:
\begin{equation}
\mathbf{V}^t_{\text{seq}} = \mathcal{E}_{t}(\mathbf{H}_{mmt} + \mathbf{E}_{pos}),\quad\mathbf{V}^s_{\text{seq}} = \mathcal{E}_{s}(\mathbf{H}_{mms} + \mathbf{E}_{spe}).
\label{eq:st_encoding}
\end{equation}

Each Transformer layer employs multi-head self-attention (MHSA) mechanisms and feed-forward networks (FFNs), incorporating residual connections and layer normalization for stable training. Global representations are derived by applying max-pooling along the sequence dimension, resulting in global temporal features $\mathbf{v}_g^t \in \mathbb{R}^{N \times D_h}$ and global spatial features $\mathbf{v}_g^s \in \mathbb{R}^{N \times D_h}$. A comprehensive global visual feature $\mathbf{v}_g \in \mathbb{R}^{N \times 2D_h}$ is formed by concatenating these two global stream representations.

\noindent\textbf{Decoupled Projection.}
To facilitate multi-grained cross-modal alignment, we propose a decoupled projection strategy. Three parallel projection networks map encoder outputs to a semantic space aligned with text embeddings $\mathbf{a} \in \mathbb{R}^{N \times D_a}$ from pre-trained language models:
\begin{equation}
\mathbf{v} = \mathcal{P}_{\text{global}}(\mathbf{v}_g), \quad \mathbf{v}_t = \mathcal{P}_{\text{temporal}}(\mathbf{v}_g^t), \quad \mathbf{v}_s = \mathcal{P}_{\text{spatial}}(\mathbf{v}_g^s),
\label{eq:decoupled_projection}
\end{equation}
where $\mathbf{v}, \mathbf{v}_t, \mathbf{v}_s \in \mathbb{R}^{N \times D_a}$, and $D_a$ denotes the dimension of the text embeddings. Each $\mathcal{P}_{*}$ represents the corresponding projection network.

\subsection{Multi-Grained Motion-Text Alignment}
To imbue the learned visual representations with semantic meaning, effective alignment with corresponding textual action labels is paramount. We achieve this through a multi-grained alignment strategy based on contrastive learning.

\noindent\textbf{Global Instance Alignment.}
The primary objective is to align the final global visual representation $\mathbf{v}$ with the corresponding action label embedding $\mathbf{a}$ at the instance level:
\begin{equation}
    \mathcal{L}_{\text{instance}} = \mathcal{L}_{\text{symm}}(\mathbf{v}, \mathbf{a}),
    \label{eq:loss_instance}
\end{equation}
where $\mathcal{L}_{\text{symm}}$ represents the symmetric contrastive loss function, ensuring the alignment objective is considered from both visual-to-text and text-to-visual perspectives:
\begin{equation}
    \mathcal{L}_{\text{symm}}(x, y) = \frac{1}{2N} \sum_{i=1}^{N} [\mathcal{L}_{C}(x^i, y^i) + \mathcal{L}_{C}(y^i, x^i)].
    \label{eq:loss_symm}
\end{equation}
The core contrastive loss component $\mathcal{L}_C$ adapted from InfoNCE is given by:
\begin{equation}
\label{eq:loss_contrastive}
\begin{split}
&\mathcal{L}_C(x^i,y^i)= \\
&-\log\frac{e^{\theta(x^i,y^i)/\tau}}{e^{\theta(x^i,y^i)/\tau} + \displaystyle\sum_{k \neq i}\bigl[e^{\theta(x^i,y^k)/\tau}
  +e^{\theta(x^i,x^k)/\tau}\bigr]},
\end{split}
\end{equation}
where $\theta(x^i, y^i) = \frac{x^{i\top} y^i}{\|x^i\| \|y^i\|}$ denotes cosine similarity, and $\tau$ is a temperature hyper-parameter. The denominator aggregates similarities over positive pairs and negative pairs within the batch, including both inter-modal $y^k$ and intra-modal $x^k$ negatives. This objective pulls together representations of corresponding visual-text pairs while pushing apart non-corresponding pairs. It further encourages the model to learn discriminative visual features by pushing apart the representations of different visual instances.

\noindent\textbf{Stream-Specific Alignment.}
To explicitly guide both the temporal and spatial processing streams towards independently capturing label-relevant information, we introduce the Stream-Specific Alignment (SSA) module. This module enforces two key alignment objectives. Firstly, a spatio-temporal alignment loss $\mathcal{L}_{ts}$ encourages the projected global temporal features $\mathbf{v}_t$ and spatial features $\mathbf{v}_s$ to individually align with the action label embedding $\mathbf{a}$:
\begin{equation}
    \mathcal{L}_{ts} = \frac{1}{2} \left( \mathcal{L}_{\text{symm}}(\mathbf{v}_t, \mathbf{a}) + \mathcal{L}_{\text{symm}}(\mathbf{v}_s, \mathbf{a}) \right).
    \label{eq:loss_ts}
\end{equation}
Furthermore, to promote coherence between the high-level semantics extracted by the two streams for the same action instance, the SSA module enforces spatio-temporal consistency via the loss $\mathcal{L}_{\text{consis}}$:
\begin{equation}
    \mathcal{L}_{\text{consis}} = \mathcal{L}_{\text{symm}}(\mathbf{v}_t, \mathbf{v}_s).
    \label{eq:loss_consis}
\end{equation}

\noindent\textbf{Fine-Grained Alignment.}
Relying solely on global features might obscure nuanced spatio-temporal cues crucial for recognizing rare actions. To capture more localized details, we propose the Fine-Grained Alignment (FGA) module. Within this module, the temporal sequence features $\mathbf{V}^t_{\text{seq}}$ are partitioned into $N_{\text{seg}}$ non-overlapping segments, and the spatial sequence features $\mathbf{V}^s_{\text{seq}}$ are grouped into $N_{\text{part}}$ semantic parts (e.g., head, arms, legs, based on joint indices). For each temporal segment $j$ and spatial part $k$, we obtain representative features $\mathbf{h}_{t,j}$ and $\mathbf{h}_{s,k}$ via max pooling within the segment/part features from $\mathbf{V}^t_{\text{seq}}$ and $\mathbf{V}^s_{\text{seq}}$, respectively. These local features are then projected using the corresponding projectors to get $\{\mathbf{v}_{t,j}\}_{j=1}^{N_{\text{seg}}}$ and $\{\mathbf{v}_{s,k}\}_{k=1}^{N_{\text{part}}}$. The fine-grained alignment loss $\mathcal{L}_{\text{part}}$ encourages these local features to align with the action label embedding $\mathbf{a}$:
\begin{equation}
\begin{split}
&\mathcal{L}_{\text{part}} = \\
&\frac{1}{2} \left( \frac{1}{N_{\text{seg}}} \sum_{j=1}^{N_{\text{seg}}} \mathcal{L}_{\text{symm}}(\mathbf{v}_{t,j}, \mathbf{a}) + \frac{1}{N_{\text{part}}} \sum_{k=1}^{N_{\text{part}}} \mathcal{L}_{\text{symm}}(\mathbf{v}_{s,k}, \mathbf{a}) \right).
    \label{eq:loss_part}
\end{split}
\end{equation}

\noindent\textbf{Training Loss.}
The final training objective function integrates these multi-grained alignment losses via a weighted sum:
\begin{equation}
    \mathcal{L}_{\text{total}} = \mathcal{L}_{\text{instance}} + \lambda_{ts}\mathcal{L}_{ts} + \lambda_{\text{consis}}\mathcal{L}_{\text{consis}} + \lambda_{\text{part}}\mathcal{L}_{\text{part}}.
    \label{eq:loss_total}
\end{equation}
Here, $\lambda_{ts}$, $\lambda_{\text{consis}}$, and $\lambda_{\text{part}}$ are hyper-parameters balancing the contribution of each component. This multi-grained optimization learns representations that preserve global action semantics while capturing discriminative spatio-temporal details, enabling robust performance across heterogeneous skeletons. During inference, the most relevant action category is identified by computing the cosine similarity $\theta$ between the global visual embedding $\mathbf{v}$ and each action label embedding $\mathbf{a}$.

\begin{table*}[t]
  \centering
  \caption{Effect of Multi-Grained Motion-Text Alignment on NTU-60 and HumanML3D. The blue numbers denote the performance improvements over the baseline model (first row). All datasets are evaluated using a single unified model.}
  \label{tab:alignment}
  \resizebox{\textwidth}{!}{
    \begin{tabular}{cccccccccc}
    \toprule
    \multicolumn{2}{c}{SSA} & FGA   & \multirow{2}{*}{Modality} & NTU-60 (3D) & NTU-60 (2D) & \multicolumn{4}{c}{HumanML3D} \\
    \cmidrule(lr){1-2} \cmidrule(lr){3-3} \cmidrule(lr){5-6} \cmidrule(lr){7-10} $\mathcal{L}_{\text{consis}}$ & $\mathcal{L}_{ts}$ & $\mathcal{L}_{\text{part}}$ &       & x-sub & x-sub & Overall & Many-shot & Medium-shot & Few-shot \\
    \midrule
          &       &       & J     & 82.09 & 86.56 & 56.76 & 67.38 & 46.74  & 42.17 \\
    \checkmark     &       &       & J     & 82.92 (\textcolor{blue}{+0.83}) & 86.70 (\textcolor{blue}{+0.14}) & 58.15 (\textcolor{blue}{+1.39}) & 68.02 (\textcolor{blue}{+0.64}) & 48.27 (\textcolor{blue}{+1.53})  & 44.72 (\textcolor{blue}{+2.55}) \\
    \checkmark     & \checkmark     &       & J     & 84.52 (\textcolor{blue}{+2.43})  & 87.46 (\textcolor{blue}{+0.90}) & 58.81 (\textcolor{blue}{+2.05}) & 69.65 (\textcolor{blue}{+2.27}) & 48.61 (\textcolor{blue}{+1.87}) & 46.90 (\textcolor{blue}{+4.73}) \\
    \checkmark     & \checkmark     & \checkmark     & J     & 84.70 (\textcolor{blue}{+2.61})  & 87.58 (\textcolor{blue}{+1.02}) & 59.34 (\textcolor{blue}{+2.58}) & 69.44 (\textcolor{blue}{+2.06}) & 49.92 (\textcolor{blue}{+3.18})  & 47.36 (\textcolor{blue}{+5.19}) \\
    \midrule
    \checkmark     & \checkmark     & \checkmark     & J+B+M & \textbf{87.01}    & \textbf{90.10}  & \textbf{61.23} & \textbf{71.51} & \textbf{52.28} & \textbf{49.36} \\
    \bottomrule
    \end{tabular}
    }
\end{table*}

\section{Experiments}
\label{sec:experiments}

\subsection{Datasets}
We utilize the representative human action datasets with varying acquisition methods, skeletal configurations, and annotation schemes:

\noindent\textbf{NTU-3D} (NTU-60~\cite{shahroudy2016ntu} \& NTU-120~\cite{liu2019ntu}): Captured from three Microsoft Kinect v2 sensors from different viewpoints, providing 25-joint 3D coordinates. NTU-60 contains 56,880 samples across 60 actions, while NTU-120 extends to 120 actions with 114,480 samples.

\noindent\textbf{NTU-2D}~\cite{duan2022revisiting}: Derived from NTU RGB+D videos using HRNet pose estimation, offering 17-joint 2D coordinates.

\noindent\textbf{HumanML3D}~\cite{guo2022generating}: One of the largest and most diverse scripted 3D motion capture (MoCap) datasets, integrating motion sequences from HumanAct12~\cite{guo2020action2motion}, AMASS~\cite{mahmood2019amass}, and others. Following SMPL~\cite{loper2015smpl} topology with 22 joints, it contains 14,616 motions paired with 44,970 text descriptions for text-to-motion generation.

\noindent\textbf{NW-UCLA}~\cite{wang2014cross}: This dataset is captured using three Microsoft Kinect v1 cameras and contains 1,494 action samples performed by 10 subjects. It includes 10 distinct action categories, with each skeleton represented by 20 3D joints.

\subsection{Evaluation Metrics}
We employ top-1 accuracy as the primary evaluation metric across all datasets. For the multi-label HumanML3D dataset, this metric assesses whether the top-scoring predicted label matches any of the ground-truth labels. To evaluate performance across varying action frequencies, we also report stratified accuracy: Head (Many-shot), Medium (Medium-shot), and Tail (Few-shot). In zero-shot learning (ZSL) experiments, we report the ZSL accuracy as well as the generalized ZSL metrics: Seen accuracy (S), Unseen accuracy (U), and Harmonic Mean (H). All datasets in each row of the table are evaluated using the same model.

\subsection{Implementation Details}
All experiments are conducted on two NVIDIA GeForce RTX 4090 GPUs. The model is trained for 400 epochs using the Adam optimizer with a batch size of 256. To improve training stability, we adopt a learning rate warm-up strategy, linearly increasing the learning rate from 0 to the peak value of $1 \times 10^{-4}$ during the first 16 epochs. The learning rate is subsequently decayed following a cosine annealing schedule. Both the temporal encoder and the spatial encoder use a Transformer with four layers. The unified skeleton structure is set to $K_{\text{unified}}=31$ and $M_{\text{unified}}=2$, determined by Eq.~(\ref{eq:unified_k}) over the training corpus. Semantic representations of action labels are obtained using the pre-trained CLIP text encoder (ViT-L/14@336px). The temperature coefficient for the contrastive loss $\mathcal{L}_C$ is set to $\tau = 0.4$. The auxiliary loss weights are set to $\lambda_{ts} = 1.0$, $\lambda_{\text{consis}} = 0.2$ and $\lambda_{\text{part}} = 0.5$. For the loss $\mathcal{L}_{\text{part}}$, we use $N_{\text{seg}} = 4$ temporal segments and $N_{\text{part}} = 4$ spatial body parts per person.

\subsection{Recognition on Heterogeneous Skeletons}

\noindent\textbf{Effect of Multi-Grained Motion-Text Alignment on Heterogeneous Skeletons.}
Table~\ref{tab:alignment} presents an ablation study evaluating the contribution of each component within our proposed multi-grained motion-text alignment framework. Starting from a baseline model trained solely with the instance-level alignment loss $\mathcal{L}_{\text{instance}}$, we progressively incorporate the other alignment losses. Improvements over the baseline are shown in parentheses. The model is jointly trained on the NTU-60 (3D and 2D versions) and HumanML3D datasets, followed by separate evaluations on each. Results show that the inclusion of the $\mathcal{L}_{\text{consis}}$, $\mathcal{L}_{ts}$, and $\mathcal{L}_{\text{part}}$ consistently enhances classification performance. This indicates that both the SSA and FGA modules facilitate the visual encoder in learning more effective representations for heterogeneous skeletons. SSA improves recognition accuracy across all evaluated datasets by promoting semantic learning through both temporal and spatial streams. Notably, the integration of FGA yields substantial performance gains in the medium-shot and few-shot categories of HumanML3D, showing its effectiveness in recognizing rare actions and mitigating bias toward head classes. When multi-modal inputs (J+B+M) are fused, the model achieves its optimal performance.

\begin{table}[t]
  \centering
  \caption{Results of action recognition on HumanML3D. All methods employ the same visual backbone for fair comparison.}
  \label{tab:humanml3d}
  \resizebox{\columnwidth}{!}{
    \begin{tabular}{lcccc}
    \toprule
    Method & Overall & Many-shot & Medium-shot & Few-shot \\
    \midrule
    RelationNet~\cite{jasani2019skeleton} & 26.05 & 54.74 & 20.69 & 5.65 \\
    CADA-VAE~\cite{schonfeld2019generalized} & 43.60  & 51.98 & 28.50  & 12.30 \\
    SMIE~\cite{zhou2023zero}  & 49.21 & 56.42 & 36.58 & 23.60 \\
    PURLS~\cite{zhu2024part} & 55.22 & 59.11 & 47.39 & 43.26 \\
    FS-VAE~\cite{wu2025frequency} & 36.60 & 55.94 & 30.05 & 13.20 \\
    TDSM~\cite{do2025bridging} & 49.95 & 51.58 & 39.57 & 24.63 \\
    \midrule
    Ours & \textbf{61.23} & \textbf{71.51} & \textbf{52.28} & \textbf{49.36} \\
    \bottomrule
    \end{tabular}}
\end{table}

\begin{table}[t]
\centering
\caption{Cross-dataset evaluation across different skeleton formats. The model is pre-trained on a single dataset with a specific skeletal format and transferred to datasets featuring different sensing modalities via linear probing, including Kinect v2 (3D), 2D pose estimations, MoCap data, and Kinect v1.}
\label{tab:cross_dataset}
\resizebox{\columnwidth}{!}{
\begin{tabular}{lllc}
\toprule
Training  & Testing  & Skeleton Format & Accuracy \\
\midrule
NTU-60 (3D) & HumanML3D      & Kinect v2 $\rightarrow$ MoCap      & 52.72 \\
NTU-60 (3D) & NTU-60 (2D)    & Kinect v2 $\rightarrow$ 2D Pose    & 75.41 \\
NTU-60 (3D) & NW-UCLA        & Kinect v2 $\rightarrow$ Kinect v1  & 91.59 \\
\midrule
HumanML3D   & NTU-60 (3D)    & MoCap $\rightarrow$ Kinect v2      & 74.81 \\
HumanML3D   & NTU-60 (2D)    & MoCap $\rightarrow$ 2D Pose        & 74.91 \\
HumanML3D   & NW-UCLA        & MoCap $\rightarrow$ Kinect v1      & 86.85 \\
\bottomrule
\end{tabular}
}
\vspace{-3mm}
\end{table}

\begin{table}[t]
  \centering
  \caption{ZSL and GZSL performance on the HumanML3D benchmark. The unseen set
  ($\mathcal{C}_{\text{novel}}$) consists of 8 tail categories; the GZSL
  candidate space covers all 400 categories. S: accuracy on seen classes.
  U: accuracy on unseen classes. H: harmonic mean. All methods use the same
  visual backbone.}
  \label{tab:zsl_humanml3d}
\resizebox{0.7\columnwidth}{!}{
    \begin{tabular}{lcccc}
    \toprule
    \multirow{2}[4]{*}{Method} & ZSL   & \multicolumn{3}{c}{GZSL} \\
\cmidrule(lr){2-2} \cmidrule(lr){3-5}          & Acc   & S     & U     & H \\
    \midrule
    RelationNet~\cite{jasani2019skeleton} & 38.60 & 23.74 & 18.91 & 21.05 \\
    CADA-VAE~\cite{schonfeld2019generalized} & 38.86 & 27.32 & 21.50 & 24.06 \\
    SMIE~\cite{zhou2023zero}  & 39.89 & 26.78 & 23.57 & 25.07 \\
    PURLS~\cite{zhu2024part}   & 40.26 & 30.92 & \textbf{34.81} & 32.75 \\
    FS-VAE~\cite{wu2025frequency}   & 36.62 & 39.24 & 21.82 & 28.04 \\
    TDSM~\cite{do2025bridging}    & 31.69 & 33.46 & 24.94 & 28.58 \\
    \midrule
    Ours  & \textbf{43.26} & \textbf{39.72} & 28.57 & \textbf{33.24} \\
    \bottomrule
    \end{tabular}%
    }
\end{table}

\begin{table}[t]
  \centering
  \caption{Performance comparison on the NW-UCLA, NTU-60 (x-sub), and NTU-120 (x-sub) datasets for heterogeneous skeleton-based action recognition. ``Train'' indicates whether the method's encoder or the fully connected (FC) classifier is trained/fine-tuned on the corresponding datasets. All results are evaluated using the pre-trained models reported in Table~\ref{tab:alignment} (last row).}
  \label{tab:ucla}
\resizebox{\columnwidth}{!}{
\begin{tabular}{lccccc}
\toprule
Method & Backbone & Train & NW-UCLA & NTU-60 & NTU-120 \\
\midrule
LongT GAN~\cite{zheng2018unsupervised} & GRU & \checkmark & 74.3 & 39.1 & 35.6 \\
P\&C~\cite{su2020predict} & GRU & \checkmark & 84.9 & 50.7 & 41.1 \\
M$\text{S}^2$L~\cite{lin2020ms2l} & GRU & \checkmark & 76.8 & 52.6 & - \\
PCRP~\cite{xu2021prototypical} & GRU & \checkmark & 87.0 & 53.9 & 41.7 \\
MCAE-MP~\cite{xu2021unsupervised} & MCAE & \checkmark & 84.9 & 65.6 & 52.8 \\
CRRL~\cite{wang2022contrast} & GRU & \checkmark & 83.8 & 67.6 & 57.0 \\
H-Transformer~\cite{cheng2021hierarchical} & Transformer & \checkmark & 83.9 & 69.3 & - \\
GL-Transformer~\cite{kim2022global} & Transformer & \checkmark & 90.4 & 76.3 & 66.0 \\
\midrule
3s-CrosSCLR~\cite{li20213d} & GCN & \checkmark & 83.6 & 77.8 & 67.9 \\
4s-ST-CL~\cite{gao2021efficient} & GCN & \checkmark & 81.2 & 68.1 & 54.2 \\
4s-MG-AL~\cite{yang2022motion} & GCN & \checkmark & 81.1 & 64.7 & 46.2 \\
CrossMoco~\cite{zeng2023contrastive} & GCN & \checkmark & 87.6 & 78.4 & - \\
3s-Colorization~\cite{yang2021skeleton} & DGCNN & \checkmark & 91.1 & 75.2 & 64.3 \\
3s-C-F Colorization~\cite{yang2023self} & DGCNN & \checkmark & 92.0 & 79.1 & 69.2 \\
\midrule
Ours & Transformer & $\times$ & 43.5 & 87.0 & - \\
Ours (w/ FC classifier) & Transformer & \checkmark & \textbf{93.5} & \textbf{87.2} & \textbf{79.5} \\
\bottomrule
\end{tabular}
}
\end{table}

\begin{figure}[tb]
    \centering
    \includegraphics[width=0.55\linewidth]{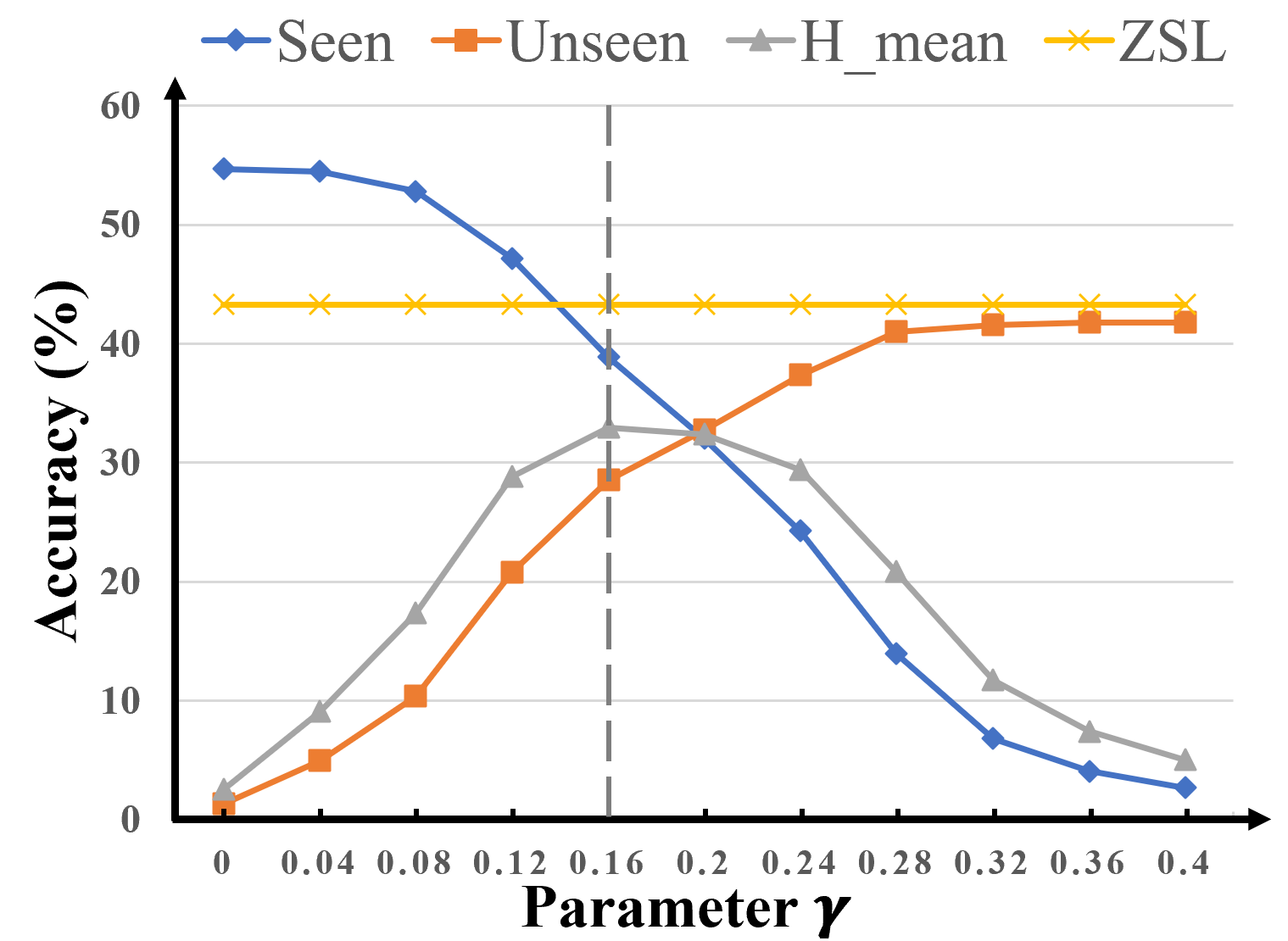}
    \caption{Sensitivity analysis of the calibration factor $\gamma$ on ZSL and GZSL performance metrics (Seen Accuracy, Unseen Accuracy, Harmonic Mean) on the HumanML3D benchmark.}
    \label{fig:zsl_humanml3d}
\end{figure}

\begin{table}[t]
  \centering
  \caption{Effect of partition strategies for Fine-Grained Alignment (FGA) on HumanML3D. Different combinations of temporal segments $N_{\text{seg}}$ and spatial partitions $N_{\text{part}}$ are evaluated to assess the influence of motion granularity on recognition performance.}
  \label{tab:partition}
  \resizebox{0.65\columnwidth}{!}{
    \begin{tabular}{cccc}
    \toprule
    $N_{\text{seg}}$ & $N_{\text{part}}$ & Medium-shot & Few-shot \\
    \midrule
    1  & 1  & 48.24  & 45.86 \\
    2  & 2  & 48.60  & 45.93 \\
    4  & 4  & \textbf{49.92} & \textbf{47.36} \\
    6  & 6  & 48.87 & 45.73 \\
    \bottomrule
    \end{tabular}}
\end{table}

\noindent\textbf{Results on the HumanML3D benchmark.}
As described in Section~\ref{sec:dataset2}, to facilitate open-vocabulary skeleton-based action recognition, we construct a benchmark based on HumanML3D, a dataset originally designed for motion synthesis. We establish representative baselines of zero-shot skeleton-based action recognition methods on this benchmark and present the results in Table~\ref{tab:humanml3d}. For a fair comparison, all baselines adopt the same visual backbone as our method. The results demonstrate that our method significantly outperforms these re-implemented baselines.

\noindent\textbf{Cross-Format Generalization Evaluation.}
To validate cross-format generalization, we train the model on a single dataset with a specific skeletal format and subsequently evaluate its transferability on other datasets with different skeleton formats. As shown in Table~\ref{tab:cross_dataset}, the model demonstrates strong generalization ability across heterogeneous skeletal representations. For example, when trained solely on HumanML3D (MoCap data), it achieves an accuracy of 74.81\% on NTU-60 (Kinect v2 data). Similarly, training on NTU-60 (3D) and testing on NW-UCLA (Kinect v1 data) yields an accuracy of 91.59\%. These results indicate that the model effectively bridges substantial structural and semantic discrepancies across skeleton types. This suggests that the unified framework learns a robust, semantically grounded representation rather than overfitting to any particular skeleton topology, validating its effectiveness in real-world heterogeneous scenarios.

\noindent\textbf{Evaluation on Heterogeneous Skeletons.}
To evaluate the model's generalization to heterogeneous skeletons and its potential for open-vocabulary action recognition, we conduct both zero-shot and transfer learning experiments using the pre-trained model. As shown in Table~\ref{tab:ucla}, we compare our method with several self-supervised learning (SSL) approaches. Under a comparable protocol where the encoder is frozen and a linear classifier is trained, our model (w/ FC classifier) achieves 93.5\% accuracy on NW-UCLA and 79.5\% on NTU-120, despite not encountering NW-UCLA data during its initial training. These results significantly outperform existing SSL baselines. In the more challenging zero-shot setting, the model performs inference directly without a classifier head and still achieves 43.5\% accuracy on NW-UCLA. This result underscores the model's capacity for open-vocabulary recognition, driven by the effective cross-domain semantic alignment.

\subsection{Open-Vocabulary Recognition}
\noindent\textbf{Zero-Shot and Generalized Zero-Shot Results.} We hold out the 8 least frequent categories of the HumanML3D benchmark as the unseen set $\mathcal{C}_{\text{novel}}$ and train on the remaining classes. Conventional ZSL restricts the candidate set to $\mathcal{C}_{\text{novel}}$, whereas generalized ZSL (GZSL) scores every sample against all 400 categories. For GZSL, candidates are ranked by the cosine similarity between $\mathbf{v}$ and $\mathbf{a}_c$, with a calibration factor $\gamma = 0.16$ subtracted from seen classes following Calibrated Stacking~\cite{chao2016empirical}. As shown in Table~\ref{tab:zsl_humanml3d}, where all baselines share the same visual backbone, our method attains 43.26\% under ZSL and a harmonic mean of 33.24\% under GZSL. This evaluation is considerably more demanding than standard NTU splits: the candidate space contains 400 rather than 5 to 24 classes, and the unseen categories are drawn from the long tail and are thus semantically infrequent. Together with the 43.5\% on the entirely unseen NW-UCLA dataset in Table~\ref{tab:ucla}, these results indicate that the multi-grained alignment transfers to novel categories without any dataset-specific classification layer.

\noindent\textbf{Influence of the Calibration Factor.}
Figure~\ref{fig:zsl_humanml3d} varies $\gamma$ on HumanML3D. At $\gamma = 0$ the model is strongly biased toward seen classes, since the contrastive objective in Eq.~(\ref{eq:loss_contrastive}) pulls visual features toward the label embeddings observed during training, a bias amplified by the 392 seen alternatives competing with only eight unseen categories. As $\gamma$ increases, Seen accuracy decreases monotonically while Unseen rises, and the harmonic mean peaks at $\gamma = 0.16$, just before the two curves cross. We adopt this value throughout. Conventional ZSL remains flat because its candidate set contains no seen classes, confirming that calibration affects only the seen--unseen trade-off rather than the underlying motion--text similarities. The decline beyond the peak indicates that calibration merely redistributes probability mass between the two label groups. Closing the remaining gap therefore requires the representation itself to place unseen categories more competitively, which we regard as the main open problem in this setting.

\subsection{Ablation Studies}

\noindent\textbf{Effect of Partition Strategies.}
The FGA module relies on partitioning temporal and spatial feature sequences to capture localized action semantics. To comprehensively assess the impact of different partition strategies, we conduct an ablation study on the HumanML3D benchmark, focusing on performance in medium-shot and few-shot categories. As shown in Table~\ref{tab:partition}, the configuration using $N_{\text{seg}} = 4$ temporal segments and $N_{\text{part}} = 4$ spatial body parts (head, arms, spine, and legs) achieves the best performance. The results highlight that the FGA module significantly improves the recognition of rare actions by leveraging fine-grained local cues. Effectively decomposing actions into meaningful spatio-temporal components contributes to improved performance under the long-tailed distribution of the HumanML3D benchmark.

\begin{table}[t]
  \centering
  \caption{Effect of different visual encoder configurations on NTU-60. Temporal-only, spatial-only, and spatio-temporal (ST) encoders with different layers are compared.}
  \label{tab:visual_encoders}
  \resizebox{0.7\columnwidth}{!}{
    \begin{tabular}{lcc}
    \toprule
    Visual encoder & NTU-60 (3D) & NTU-60 (2D) \\
    \midrule
    Temporal & 82.93 & 86.51 \\
    Spatial & 80.86 & 80.03 \\
    ST (2L) & 83.35 & 86.77 \\
    ST (4L) & \textbf{84.70} & \textbf{87.58} \\
    \bottomrule
    \end{tabular}}
\end{table}

\noindent\textbf{Effect of Different Encoders.}
Table~\ref{tab:visual_encoders} compares temporal-only, spatial-only, and
spatio-temporal (ST) encoders with varying depths. ST encoders
consistently outperform single-stream ones, and increasing the depth from
two to four Transformer layers brings further gains. The framework is
also robust to the choice of pre-trained text encoders, with MPNet and
CLIP variants yielding comparable accuracy (see the supplementary
material).

\begin{table}[t]
\centering
\caption{Ablation study on unified representation strategies. We compare three filling methods: (1) zero padding, (2) learnable padding, where the missing joints are parameterized, and (3) kinematic imputation, which synthesizes missing joints from semantically adjacent ones.
}
\label{tab:unified_representation}
\resizebox{\linewidth}{!}{
\begin{tabular}{l|cc|cccc}
\toprule
\multirow{2}{*}{Padding Method} & NTU-60 & NTU-60 & \multicolumn{4}{c}{HumanML3D} \\
& (3D) & (2D) & Overall & Many-shot & Medium-shot & Few-shot \\
\midrule
Zero Padding & 85.87 & \textbf{90.46}& 60.77 & 70.65 & 51.72 & 48.22 \\
Learnable Padding & 86.32 & 88.76 & 60.89 & 71.21 & 51.39 & 48.72 \\
Kinematic Imputation & \textbf{87.01} & 90.10 & \textbf{61.23} & \textbf{71.51} & \textbf{52.28} & \textbf{49.36} \\
\bottomrule
\end{tabular}
}
\end{table}

\begin{figure}[tb]
\centering
\subfloat[Temperature $\tau$]{
    \includegraphics[width=0.45\linewidth]{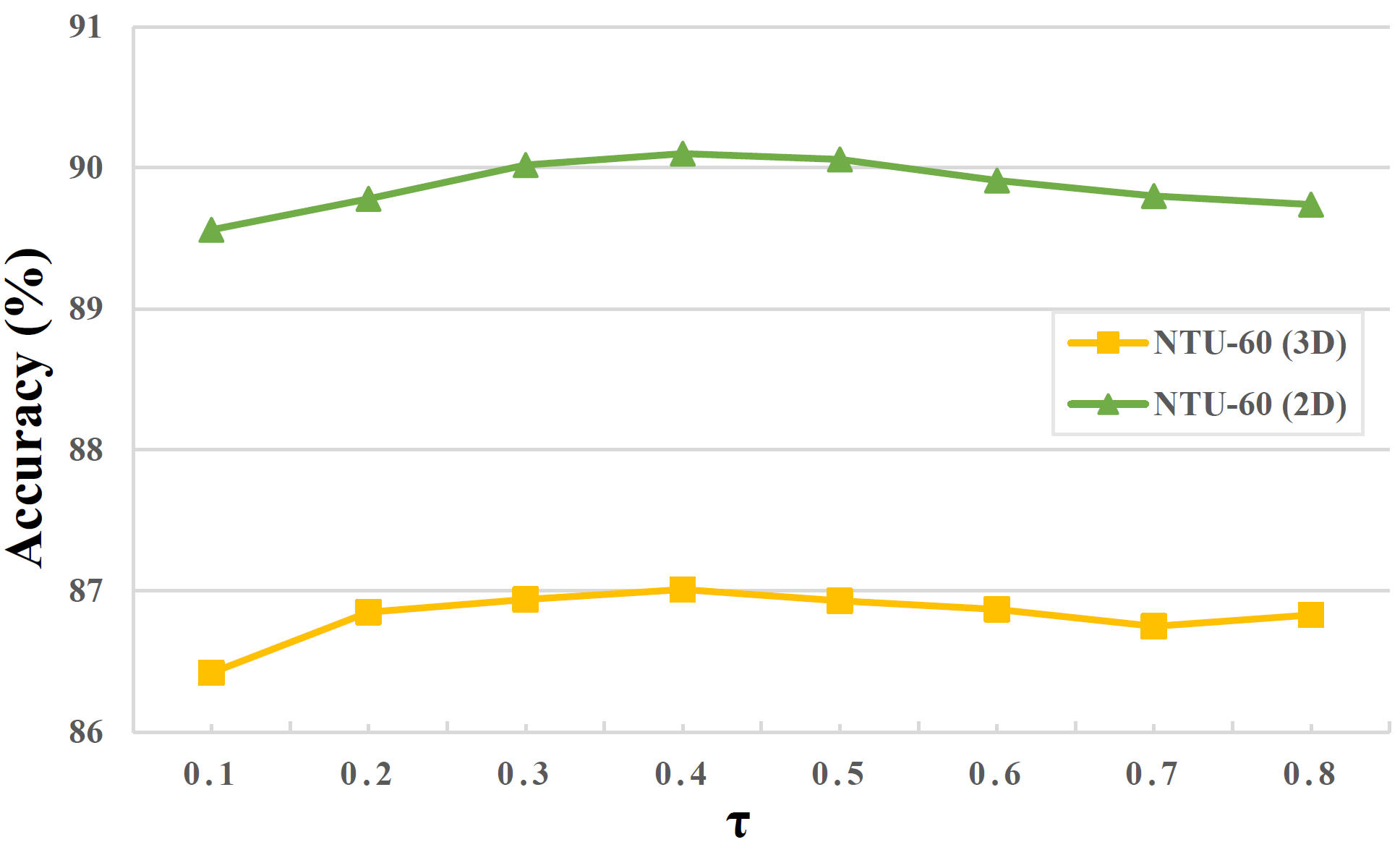}
}
\hfil
\subfloat[Loss weight $\lambda_{ts}$]{
    \includegraphics[width=0.45\linewidth]{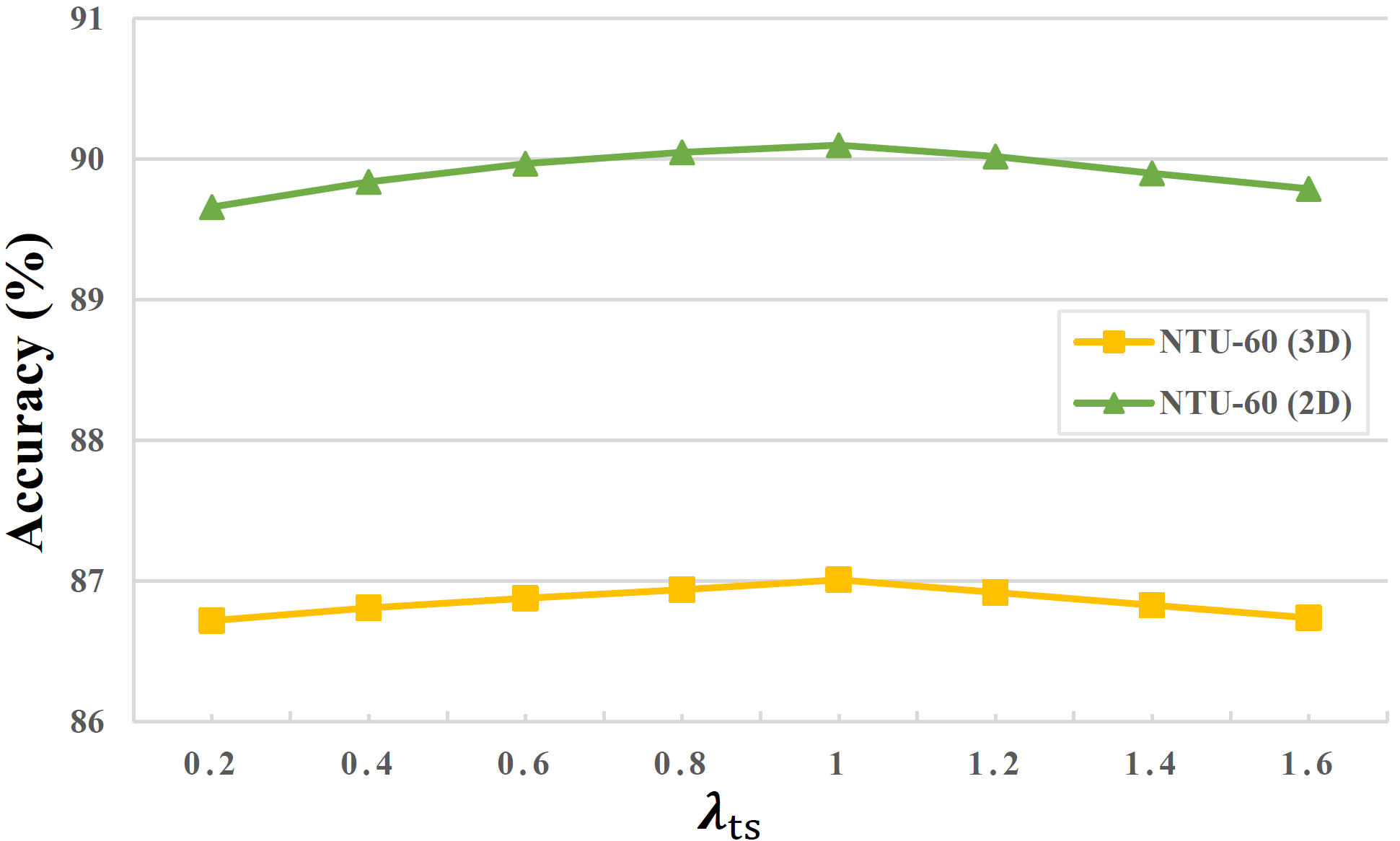}
}

\subfloat[Loss weight $\lambda_{\text{consis}}$]{
    \includegraphics[width=0.45\linewidth]{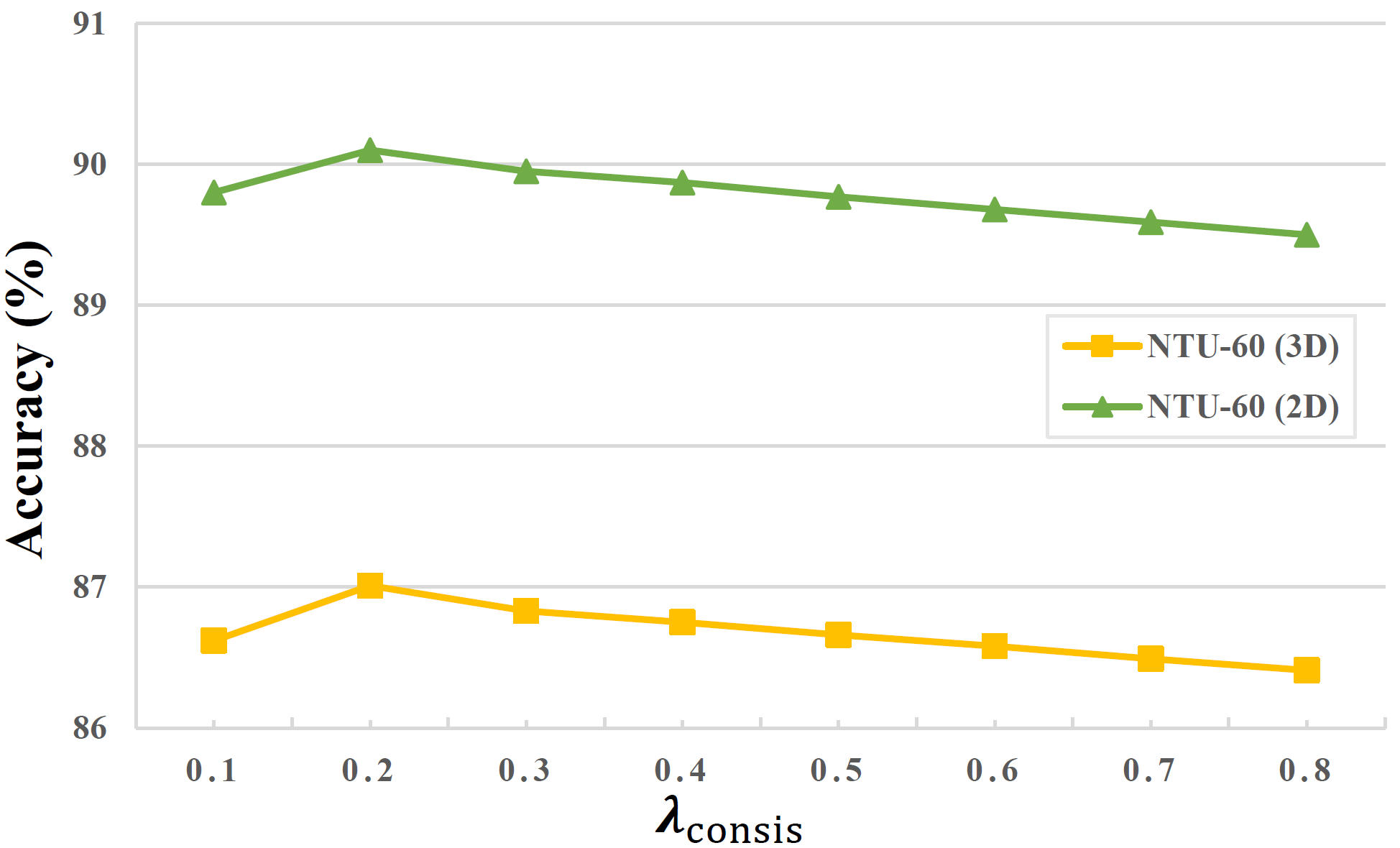}
}
\hfil
\subfloat[Loss weight $\lambda_{\text{part}}$]{
    \includegraphics[width=0.45\linewidth]{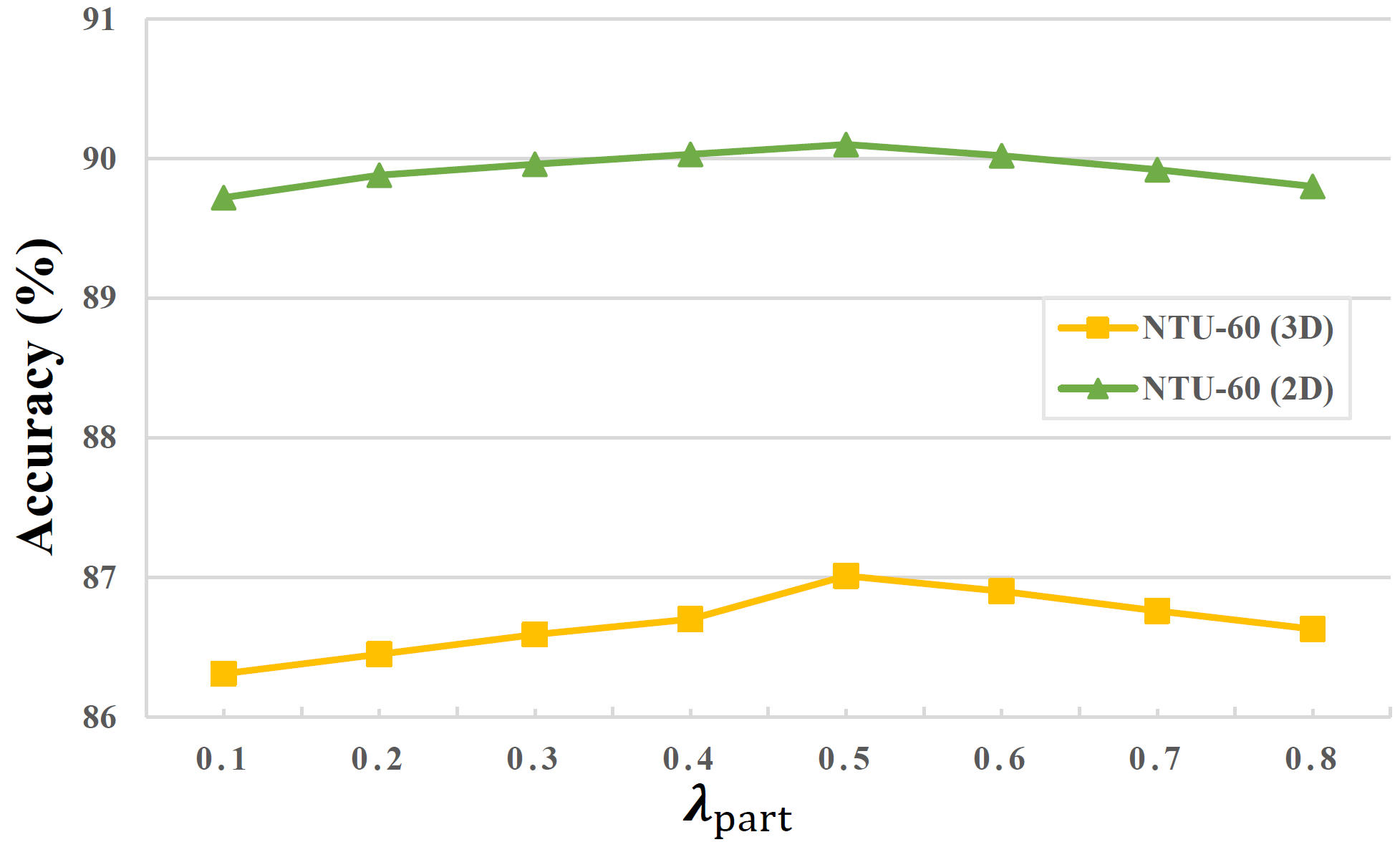}
}
\caption{Sensitivity analysis of the hyper-parameters.}
\label{fig:hyperparameter_sensitivity}
\end{figure}

\noindent\textbf{Ablation on Unified Skeleton Representation.}  
The unified skeleton representation is the basic step through which the framework handles heterogeneous inputs. As shown in Table~\ref{tab:unified_representation}, kinematic imputation achieves favorable performance. While maintaining this performance, kinematic imputation provides a skeleton unification scheme that is more anatomically plausible and introduces no additional trainable parameters. Overall, kinematic imputation performs most consistently across heterogeneous skeleton formats and is therefore adopted as the default unified representation strategy.

\noindent\textbf{Hyper-parameter Sensitivity Analysis.}
Figure~\ref{fig:hyperparameter_sensitivity} varies each hyper-parameter
of the alignment loss on NTU-60 while keeping the others at their optimal
values. Performance remains stable across a broad and practical range in
all cases. The temperature $\tau$ peaks at 0.4, $\lambda_{ts}$ around
1.0, and $\lambda_{\text{part}}$ at 0.5. Notably,
$\lambda_{\text{consis}}$ is most effective at a small value of 0.2,
suggesting that forcing excessive similarity between the temporal and
spatial streams hinders the learning of complementary features. We adopt
this configuration throughout.

\begin{figure}[tb]
    \centering
    \includegraphics[width=0.7\linewidth]{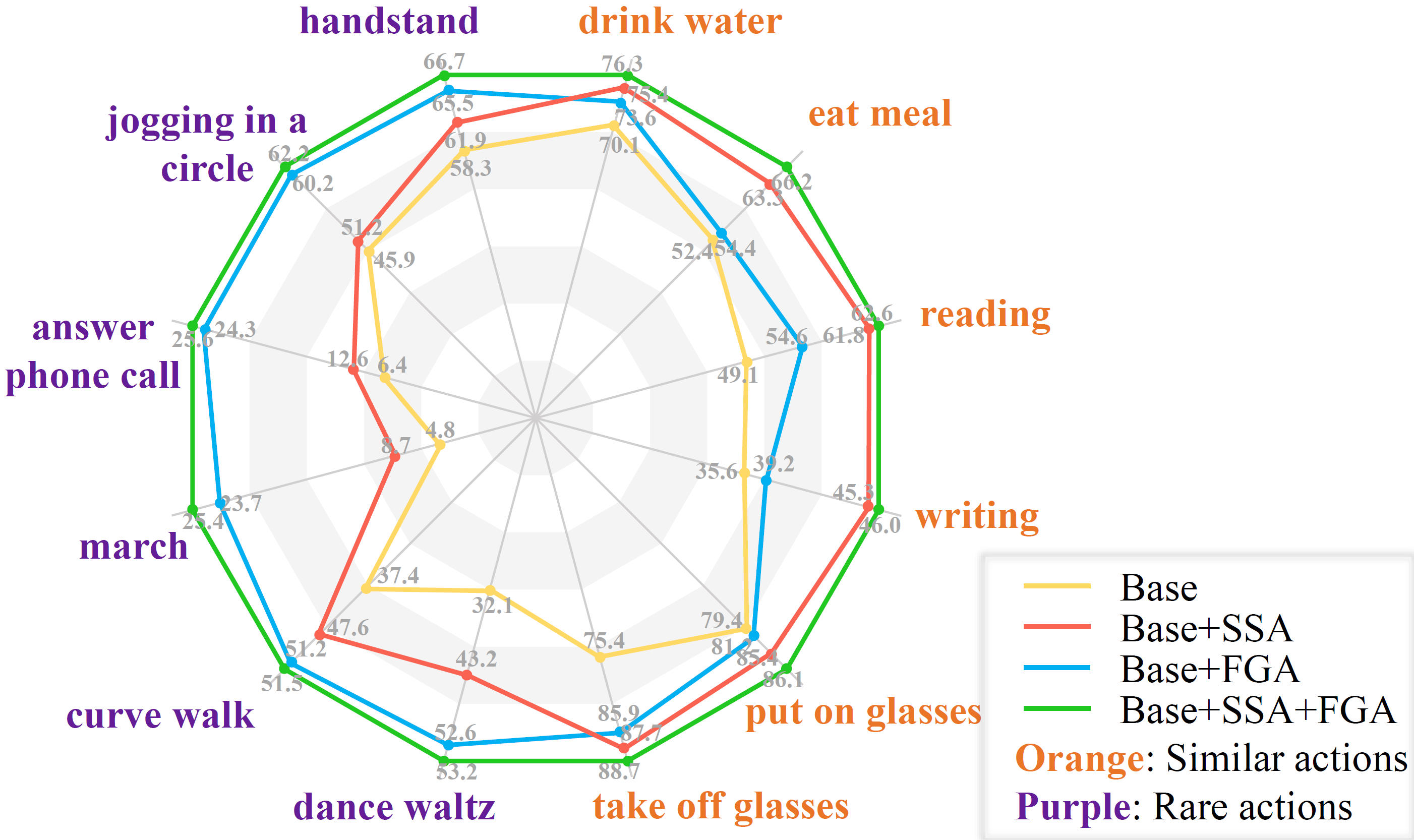}
    \caption{Radar chart of per-class accuracy improvements from SSA and FGA modules, highlighting NTU-60 similar actions (orange) and HumanML3D rare actions (purple).}
    \label{fig:radar}
\end{figure}

\noindent\textbf{Influence of Alignment Modules.}
Figure~\ref{fig:radar} reports per-class accuracy on representative categories
as SSA and FGA are progressively added to the instance-only baseline. SSA
(red) improves most categories, with the clearest gains on semantically
confusable NTU-60 pairs such as ``reading'' and ``writing'' (orange),
indicating that independent stream alignment yields more discriminative
global features. FGA (blue) is most effective on rare HumanML3D categories
such as ``handstand'' and ``jogging in a circle'' (purple), where localized
temporal segments and body parts supply cues that global alignment alone
overlooks. The combined model (green) performs best, confirming that the two
modules are complementary.

\section{Conclusion}
\label{sec:conclusion}
In this work, we study the novel problem of heterogeneous skeleton-based action recognition with open vocabularies, taking an important step toward the application of skeleton-based action recognition in real-world scenarios. We construct a large-scale dataset named Heterogeneous Open-Vocabulary (HOV) Skeleton to accommodate this task. We propose a Transformer-based model with unified skeleton representation and multi-grained motion-text alignment to achieve an all-in-one solution for skeleton-based action recognition. Extensive experiments demonstrate that the proposed model not only excels at recognizing human actions from heterogeneous data sources but also exhibits superior transferability across diverse scenarios in skeleton-based action recognition.

\noindent\textbf{Limitations.} Direct transfer of the pre-trained model to unseen scenarios remains less effective than scenario-specific fine-tuning, highlighting the need for more robust and scalable open-vocabulary text--skeleton pre-training. Moreover, although the unified representation can in principle accommodate non-human articulated structures, our evaluation is limited to human skeletons, and extending the framework to humanoid robot kinematic data remains future work.

\bibliographystyle{IEEEtran}
\bibliography{main}

\twocolumn[{
\centering
{\LARGE\bfseries Supplementary Material for\\[4pt]
``Toward Universal Skeleton-Based Action Recognition across
Heterogeneous Skeletons and Open Vocabularies''\par}
\vspace{1em}
}]

\appendices
\setcounter{table}{0}
\setcounter{figure}{0}
\setcounter{equation}{0}
\renewcommand{\thetable}{S\Roman{table}}
\renewcommand{\thefigure}{S\arabic{figure}}
\renewcommand{\theequation}{S\arabic{equation}}

\section{Unified Skeleton Representation Details}
\label{appendix:unified_representation}

\begin{figure*}[t]
    \centering
    \includegraphics[width=\textwidth]{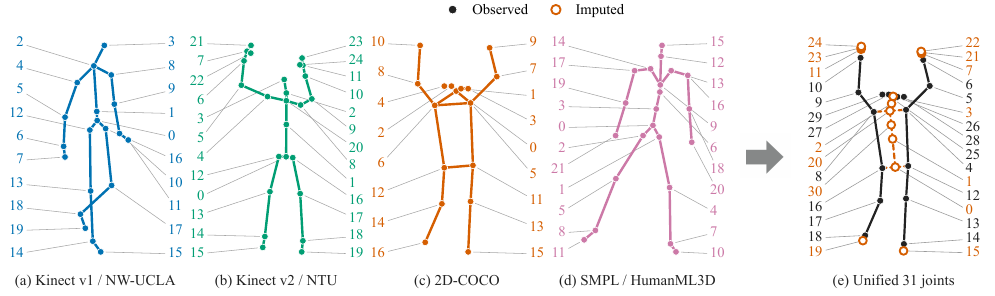}
    \caption{Unification of heterogeneous skeletons. (a)--(d)~Real samples
    from the four source formats with their native joint indexing.
    (e)~The 2D pose sample from~(c) after remapping and imputation in the
    unified 31-joint space, where solid dots denote observed joints and
    hollow circles denote imputed joints. The 2D pose format is shown
    because it requires the most imputation among the four formats.}
    \label{fig:unified_appendix}
\end{figure*}

\noindent\textbf{Joint correspondence.}
Table~\ref{tab:joint_correspondence} lists the mapping from each source format to the unified index space of $K_{\text{unified}} = 31$ joints shown in Figure~\ref{fig:unified_appendix}(e). The Kinect v2 format populates indices 0 to 24 and serves as the base layout. The 2D pose format additionally provides five facial landmarks (indices 25 to 29), and the MoCap format provides an upper-spine joint (index 30). Joints that a source does not provide are marked as absent and imputed by the rules below. Observed joints are never modified. Figure~\ref{fig:unified_appendix} visualizes the correspondence with real samples from each format.

\noindent\textbf{Missing-joint imputation.}
Absent joints are inferred from observed ones through three operators. \textit{Interpolation} recovers trunk joints as midpoints of observed pairs. The pelvis is placed at the midpoint of the two hips, the shoulder center at the midpoint of the two shoulders, the spine and upper-spine joints at midpoints along the trunk, and the neck at the midpoint of the shoulder center and the head. For 2D skeletons, the head is recovered as the mean of the observed facial landmarks. \textit{Extrapolation} recovers terminal joints by extending along the kinematic chain,
\begin{equation}
    \mathbf{p}_{\text{target}} = \mathbf{p}_{\text{anchor}} +
    \rho\,(\mathbf{p}_{\text{anchor}} - \mathbf{p}_{\text{parent}}),
    \label{eq:extrapolation}
\end{equation}
with $\rho = 0.25$ for hands along the elbow-to-wrist direction, $\rho = 0.15$ for feet along the knee-to-ankle direction, and $\rho = 0.18$ and $0.32$ for hand tips and thumbs along the wrist-to-hand direction. \textit{Proxy assignment} anchors facial landmarks absent from 3D sources at the head joint. The rules are applied in a fixed order so that later rules can build on earlier imputations. For 2D pose skeletons, for instance, the pelvis and shoulder center are first interpolated from the hips and shoulders, after which the spine and neck are placed along the reconstructed trunk and the hand tips are extrapolated from the reconstructed hands. The amount of imputation varies across formats. Kinect v2 skeletons require only the facial landmarks and the upper-spine joint, whereas 2D pose skeletons additionally require the trunk, head, hands, and feet.

\noindent\textbf{Coordinate dimensionality.}
All sources are mapped to $C_{\text{unified}} = 3$. For 2D sources the depth channel is set to zero, so that the embedding module receives a fixed input width regardless of the source, and 2D and 3D skeletons share the same encoder without architectural changes.

\noindent\textbf{Member dimension.}
Sequences containing fewer than $M_{\text{unified}}$ individuals are padded by replicating the observed person.

\section{Detailed Dataset Construction and Rationale}
\label{appendix:benchmark_construction}

This section provides a detailed rationale for the construction of our HOV Skeleton dataset, particularly the processing of the HumanML3D benchmark. We address the comparison with related datasets, and the sensitivity analysis for our clustering choices.

\begin{table}[t]
\centering
\caption{Joint correspondence between each source format and the unified 31-joint representation shown in Figure~\ref{fig:unified_appendix}(e). ``--'' indicates that the joint is absent from the source and is imputed by the rules in Appendix~\ref{appendix:unified_representation}.}
\label{tab:joint_correspondence}
\resizebox{0.95\columnwidth}{!}{
\begin{tabular}{clcccc}
\toprule
Unified & Joint name & Kinect v1 & Kinect v2 & 2D pose & MoCap \\
\midrule
0  & pelvis          & 0  & 0  & -- & 0  \\
1  & spine           & 1  & 1  & -- & 3  \\
2  & neck            & -- & 2  & -- & 12 \\
3  & head            & 3  & 3  & -- & 15 \\
4  & left shoulder   & 8  & 4  & 5  & 13 \\
5  & left elbow      & 9  & 5  & 7  & 16 \\
6  & left wrist      & 10 & 6  & 9  & 18 \\
7  & left hand       & 11 & 7  & -- & 20 \\
8  & right shoulder  & 4  & 8  & 6  & 14 \\
9  & right elbow     & 5  & 9  & 8  & 17 \\
10 & right wrist     & 6  & 10 & 10 & 19 \\
11 & right hand      & 7  & 11 & -- & 21 \\
12 & left hip        & 16 & 12 & 11 & 1  \\
13 & left knee       & 17 & 13 & 13 & 4  \\
14 & left ankle      & 18 & 14 & 15 & 7  \\
15 & left foot       & 19 & 15 & -- & 10 \\
16 & right hip       & 12 & 16 & 12 & 2  \\
17 & right knee      & 13 & 17 & 14 & 5  \\
18 & right ankle     & 14 & 18 & 16 & 8  \\
19 & right foot      & 15 & 19 & -- & 11 \\
20 & shoulder center & 2  & 20 & -- & 9  \\
21 & left hand tip   & -- & 21 & -- & -- \\
22 & left thumb      & -- & 22 & -- & -- \\
23 & right hand tip  & -- & 23 & -- & -- \\
24 & right thumb     & -- & 24 & -- & -- \\
25 & nose            & -- & -- & 0  & -- \\
26 & left eye        & -- & -- & 1  & -- \\
27 & right eye       & -- & -- & 2  & -- \\
28 & left ear        & -- & -- & 3  & -- \\
29 & right ear       & -- & -- & 4  & -- \\
30 & upper spine     & -- & -- & -- & 6  \\
\bottomrule
\end{tabular}}
\end{table}

\subsection{Comparison with the BABEL Dataset}
Although both our processed HumanML3D benchmark and the BABEL dataset~\cite{punnakkal2021babel} are derived from AMASS~\cite{mahmood2019amass}, they differ substantially in design philosophy and suitability for universal action recognition. The major distinctions are summarized in Table~\ref{tab:babel_comparison} and elaborated below.

\noindent\textbf{Full Sequences vs. Short Clips.}
This represents the most fundamental difference. BABEL provides frame-level annotations for short motion clips, which often lack sufficient temporal context to support robust representation learning. As a result, performance is constrained: our re-implementation achieves 42.45\% accuracy on BABEL's single-label task, closely matching its official 41.14\%, underscoring the difficulty of further improvement. In contrast, our benchmark uses full action sequences, enabling models to exploit long-range temporal dynamics, learn richer features, and achieve higher recognition accuracy.

\noindent\textbf{Multi-Label vs. Single-Label.}
AMASS sequences frequently contain multiple co-occurring actions. While BABEL simplifies these into a single-label formulation, our benchmark preserves this complexity by adopting a multi-label setting, which more accurately reflects the intrinsic structure of the underlying motion data.

\noindent\textbf{Preservation of Heterogeneity.}
To support research on heterogeneous skeletons, our benchmark retains the SMPL joint structure provided by AMASS. In contrast, publicly released BABEL data converts all motions into a unified Kinect v2 skeleton, removing the structural diversity that is central to our investigation.

\noindent\textbf{Comprehensive Evaluation.}
Our benchmark is specifically designed to facilitate rigorous analysis of generalization performance by organizing the labels into a long-tailed distribution (Many/Medium/Few). This evaluation perspective is not a primary focus of the BABEL benchmark.

In summary, the HumanML3D benchmark offers a more suitable data representation and task formulation, preserves essential structural heterogeneity, and enables long-tail performance assessment, making it a more robust and realistic testbed for universal action recognition.

\subsection{Sensitivity to Number of Label Clusters (K)}
The choice of the number of clusters (K) in our label space clustering step is a critical hyper-parameter. To justify our choice of K = 400, we conducted an empirical study by training and evaluating our model with different values of K. The results are presented in Table~\ref{tab:sensitivity_k}.

\begin{table}[t]
\centering
\caption{Comparison between our processed HumanML3D benchmark and the BABEL dataset.}
\label{tab:babel_comparison}
\resizebox{\columnwidth}{!}{
\begin{tabular}{l|l|l}
\toprule
Attribute & BABEL Dataset & Our HumanML3D Benchmark \\
\midrule
Data Unit & Short motion clips & Complete action sequences \\
Task Formulation & Single-Label Classification & Multi-Label Classification \\
Skeleton Format & Uniform (Kinect v2) & Original SMPL format \\
Evaluation Focus & General classification & Long-tail distribution analysis \\
Reported Accuracy & 42.45\% (Top-1, official 41.14\%) & 61.23\% (Overall, multi-label) \\
\bottomrule
\end{tabular}
}
\end{table}

\begin{table}[t]
\centering
\caption{Sensitivity analysis for the number of clusters (K) on the HumanML3D benchmark. Results show that K=400 provides the best trade-off.}
\label{tab:sensitivity_k}
\resizebox{0.9\columnwidth}{!}{
\begin{tabular}{c|cccc}
\toprule
K & Overall & Many-shot & Medium-shot & Few-shot \\
\midrule
200 & 60.87 & 70.15 & 51.33 & \textbf{49.91} \\
\textbf{400} & \textbf{61.23} & \textbf{71.51} & \textbf{52.28} & 49.36 \\
600 & 60.54 & 70.89 & 51.02 & 48.03 \\
\bottomrule
\end{tabular}
}
\end{table}

As the results demonstrate, the model's performance is stable across the tested range of K, indicating the robustness of our overall framework to this choice. However, K=400 achieves the best performance across most metrics. A smaller K (200) leads to a slight performance drop, likely because distinct semantic concepts are merged, reducing the label space's expressiveness. A larger K (600) also results in a minor decline, which we attribute to the creation of an overly sparse long tail, making it harder for the model to learn robust representations for the rarest classes. Therefore, our choice of K=400 is empirically justified as providing the optimal balance between semantic richness and a learnable class distribution.

\subsection{HumanML3D Annotation Processing}
\label{sec:appendix_annotations}

To provide a concrete illustration of our data processing pipeline for the HumanML3D benchmark, this section details the transformation from raw, descriptive text into structured, multi-label annotations. The original annotations in HumanML3D are free-form natural language sentences, which are rich in detail but unsuitable for direct use as classification labels due to their narrative style and semantic ambiguity. Our annotation pipeline is designed to address this. First, we employ a Large Language Model (LLM) to parse each sentence and extract the core atomic action verbs or short phrases. This step filters out stylistic modifiers, narrative context, and redundant information. Second, a clustering algorithm consolidates the vast vocabulary of extracted terms into a coherent set of 400 semantic action classes, merging synonyms and morphological variations (e.g., `jumped', `jumps', and `jumping' are unified into a single `jump' class).

Table~\ref{tab:annotation_examples} showcases several examples of this transformation. It contrasts the original descriptive text with the final, structured multi-label action set used to train our model. The resulting labels are more atomic, consistent, and directly suitable for a multi-label classification framework.

\begin{table*}[t]
\centering
\caption{Examples of the annotation transformation process on the HumanML3D benchmark. Original free-form text descriptions are converted into a structured set of multi-label action classes.}
\label{tab:annotation_examples}
\begin{tabular}{ll}
\toprule
Original Text Description & Processed Action Labels \\
\midrule
a person is walking, then crouches and crawls forward. & [`walk', `crouch', `crawl'] \\
someone is waving both their hands in the air. & [`wave hands'] \\
the person takes a few steps and then begins to skip. & [`walk', `skip'] \\
a person is punching the air with their left and right fists. & [`punch'] \\
the person is running in a circle. & [`run in circle'] \\
a person hops on one leg and then switches to the other. & [`hop', `switch leg'] \\
a person bends down to tie their shoe. & [`bend down', `tie shoe'] \\
a person turns 360 degrees and then claps. & [`turn', `clap'] \\
the person is marching in place while pumping their arms. & [`march', `pump arms'] \\
a person is looking left, then looking right. & [`look left', `look right'] \\
the person kicks a ball and then runs after it. & [`kick', `run'] \\
a person stretches their arms out to their sides. & [`stretch arms'] \\
\bottomrule
\end{tabular}
\end{table*}

\section{Additional Results}
\label{appendix:additional_results}

\subsection{Recognition on NTU RGB+D}
Table~\ref{tab:ntu_comparison} presents a comparative analysis of our method against state-of-the-art supervised and unsupervised techniques on the NTU RGB+D 60 (NTU-60) and NTU RGB+D 120 (NTU-120) datasets. The results reported for our method utilize the multi-modal input strategy (Joint+Bone+Motion), which empirically yielded the best performance. It is important to note that while our training paradigm incorporates action label semantics (hence, ``supervised'' in a broad sense), our architecture is distinct from traditional supervised methods as it does not employ a final classification head. Instead, recognition is achieved through aligning visual features with textual action label embeddings, a design choice aimed at facilitating open-vocabulary capabilities. As shown in Table~\ref{tab:ntu_comparison}, our approach achieves competitive performance, particularly on the NTU-60 dataset. ``Ours (2D)'' attains 90.1\% on NTU-60 x-sub and 96.6\% on NTU-60 X-View. These results are comparable to several fully supervised methods that utilize dedicated classification layers. This strong performance, achieved without a dataset-specific classifier and with a focus on open-set generalizability, underscores the efficacy of our multi-grained motion-text alignment in learning discriminative and robust representations. Furthermore, our method demonstrates a notable advantage in computational efficiency. Due to the early fusion strategy for multi-modal inputs, our approach achieves a relatively low computational complexity of 3.46 GFLOPs. This efficiency, combined with strong performance, highlights the practical viability of our framework.

\begin{table*}[t]
  \centering
  \caption{Comparative performance on NTU-60 and NTU-120. Our method performs recognition on both the 2D and 3D versions of NTU through visual-text alignment, without relying on a dedicated classification head. The results presented are the best multi-modal performances of each method.}
  \label{tab:ntu_comparison}
\resizebox{0.7\linewidth}{!}{
    \begin{tabular}{lcccccc}
    \toprule
    \multirow{2}[4]{*}{Method} & \multirow{2}[4]{*}{Publication} & \multirow{2}[4]{*}{FLOPs/G} & \multicolumn{2}{c}{NTU-60} & \multicolumn{2}{c}{NTU-120} \\
\cmidrule{4-7}          &       &       & x-sub & x-view & x-sub & x-set \\
    \midrule
    \textbf{\textit{Supervised Methods}} &       &       &       &       &       &  \\
    SkeletonGCL~\cite{huang2023graph} & ICLR 2023 & -     & 92.8  & 97.1  & 89.8  & 91.2 \\
    FR-Head~\cite{zhou2023learning} & CVPR 2023 & -     & 92.8  & 96.8  & 89.5  & 90.9 \\
    GAP~\cite{xiang2023generative}   & ICCV 2023 & -     & 92.9  & 97.0    & 89.9  & 91.1 \\
    HD-GCN~\cite{lee2023hierarchically} & ICCV 2023 & 7.08   & 93.4  & 97.2  & 90.1  & 91.6 \\
    JT-GraphFormer~\cite{zheng2024spatio} & AAAI 2024 & -     & 93.4  & 97.5  & 89.9  & 91.7 \\
    DS-GCN~\cite{xie2024dynamic} & AAAI 2024 & -     & 93.1  & 97.5  & 89.2  & 91.1 \\
    BlockGCN~\cite{zhou2024blockgcn} & CVPR 2024 & 6.52  & 93.1  & 97.0    & 90.3  & 91.5 \\
    NDMHIC~\cite{chen2025skeleton} & AAAI 2025 & -     & 93.7  & 97.3  & 90.6  & 91.7 \\
    VA-AR~\cite{wei2025va} & AAAI 2025 & -     & 93.1  & 97.2  & 90.3  & 91.5 \\
    ProtoGCN~\cite{liu2025revealing} & CVPR 2025 & -     & 93.8  & 97.8  & 90.9  & 92.2 \\
    \midrule
    \textbf{\textit{Unsupervised Methods}} &       &       &       &       &       &  \\
    AimCLR~\cite{guo2022contrastive} & AAAI 2022 & 3.45  & 78.9  & 83.8  & 68.2  & 68.8 \\
    CPM~\cite{zhang2022contrastive}   & ECCV 2022 & 6.66  & 83.2  & 87.0    & 73.0    & 74.0 \\
    CMD~\cite{mao2022cmd}   & ECCV 2022 & 17.28 & 84.1  & 90.9  & 74.7  & 76.1 \\
    HiCLR~\cite{zhang2023hierarchical} & AAAI 2023 & 7.08  & 78.8  & 83.1  & 67.3  & 69.9 \\
    PSTL~\cite{zhou2023self}  & AAAI 2023 & 3.45  & 79.1  & 83.8  & 69.2  & 70.3 \\
    HYSP~\cite{francohyperbolic}  & ICLR 2023 & 72.72     & 79.1  & 85.2  & 64.5  & 67.3 \\
    ActCLR~\cite{lin2023actionlet} & CVPR 2023 & 8.98     & 84.3  & 88.8  & 74.3  & 75.7 \\
    RVTCLR~\cite{zhu2023modeling} & ICCV 2023 & -     & 79.7  & 84.6  & 68    & 68.9 \\
    UmURL~\cite{sun2023unified} & ACM MM 2023 & 5.22  & 84.4  & 91.4  & 75.9  & 77.2 \\
    RMDRP~\cite{gong2025rethinking} & AAAI 2025 &  -     & 86.9  & 91.0    & 80.0    & 81.5 \\
    \midrule
    Ours (3D) & -     & 3.46  & 87.0 & 93.2 & 79.1 & 82.2 \\
    Ours (2D) & -     & 3.46  & 90.1  & 96.6 & 81.3 & 84.5 \\
    \bottomrule
    \end{tabular}%
    }
\end{table*}

\subsection{Cross-Dataset Generalization to NTU-120}
To further investigate the transferability and generalization capabilities of the representations learned by our model, particularly when trained on datasets with different skeletal structures and action vocabularies, we conduct a cross-dataset evaluation on the NTU-120 benchmark. Table~\ref{tab:cross_dataset_ntu120} compares the performance of our method under different training regimes with UmURL~\cite{sun2023unified}. We consider two variants of our approach: one pre-trained solely on HumanML3D (``Ours (HumanML3D)''), and another pre-trained on a combination of NTU-60 and HumanML3D (``Ours (NTU60+HumanML3D)''). During evaluation, we freeze the pre-trained backbone and train a linear classifier on the NTU-120 dataset. The results indicate that our model achieves substantial cross-dataset transfer performance on NTU-120 even when pre-trained only on HumanML3D, which differs significantly from NTU-120 in both skeleton structure (SMPL-based with 22 joints vs. Kinect v2-based with 25 joints) and action semantics. Furthermore, incorporating NTU-60 into pre-training (``Ours (NTU60+HumanML3D)'') significantly improves generalization to NTU-120. With multi-modal input (J+B+M), our model achieves the best performance among all compared methods. These findings further highlight the effectiveness of our method in learning generalizable representations for heterogeneous action recognition.

\begin{table}[t]
  \centering
  \caption{Cross-dataset generalization performance on NTU-120. ``Ours (HumanML3D)'' indicates pre-trained solely on HumanML3D, while ``Ours (NTU60+HumanML3D)'' indicates pre-trained on the combined datasets, both followed by linear classification on NTU-120.}
  \label{tab:cross_dataset_ntu120}
\resizebox{0.9\columnwidth}{!}{
    \begin{tabular}{lccc}
    \toprule
    \multirow{2}[4]{*}{Method} & \multirow{2}[4]{*}{Modality} & \multicolumn{2}{c}{NTU-120} \\
\cmidrule{3-4}          &       & x-sub & x-set \\
    \midrule
    UmURL~\cite{sun2023unified} & J     & 73.5  & 74.3 \\
    Ours (HumanML3D) & J     & 57.8 & 58.0 \\
    Ours (NTU60+HumanML3D) & J     & 76.3 & 82.9 \\
    \midrule
    UmURL~\cite{sun2023unified} & J+B+M & 75.9  & 77.2 \\
    Ours (HumanML3D) & J+B+M & 63.8 & 64.9 \\
    Ours (NTU60+HumanML3D) & J+B+M & \textbf{79.5} & \textbf{86.1} \\
    \bottomrule
    \end{tabular}%
    }
\end{table}

\begin{table}[t]
\centering
\caption{Zero-shot action recognition performance on NTU-60 and NTU-120. Our end-to-end model achieves competitive results compared to two-stage ZSAR methods.}
\label{tab:zsl_ntu}
\resizebox{\columnwidth}{!}{
\begin{tabular}{lccccc}
\toprule
\multirow{2}{*}{Method} & \multirow{2}{*}{Venue} & \multicolumn{2}{c}{NTU-60 (\%)} & \multicolumn{2}{c}{NTU-120 (\%)} \\
\cmidrule{3-6}  &  & 55/5 & 48/12 & 110/10 & 96/24 \\
\midrule
DeViSE~\cite{frome2013devise}        & NeurIPS 2013 & 60.72 & 24.51 & 47.49 & 25.74 \\
RelationNet~\cite{jasani2019skeleton}   & ICCV 2019 & 40.12 & 30.06 & 52.59 & 29.06 \\
ReViSE~\cite{hubert2017learning}        & ICCV 2017 & 59.31 & 17.49 & 55.04 & 32.89 \\
JPoSE~\cite{wray2019fine}         & ICCV 2019 & 64.82 & 28.75 & 51.93 & 32.44 \\
CADA-VAE~\cite{schonfeld2019generalized}      & CVPR 2019 & 76.84 & 28.96 & 59.34 & 33.57 \\
SynSE~\cite{gupta2021syntactically}         & ICIP 2021 & 75.81 & 33.30 & 62.69 & 38.70 \\
SMIE~\cite{zhou2023zero}          & ACM MM 2023 & 77.98 & 40.18 & 65.74 & 45.30 \\
SA-DVAE~\cite{li2024sa}     & ECCV 2024 & 82.37 & 41.38 & 68.77 & 46.12 \\
STAR~\cite{chen2024fine}        & ACM MM 2024 & 81.40 & 45.10 & 63.30 & 44.30 \\
GZSSAR~\cite{li2023multi}   & ICIG 2023 & 83.30 & 49.80 & 72.00 & 60.70 \\
InfoCPL~\cite{xu2025information}  & TMM 2025 & 85.91 & 53.32 & \textbf{74.81} & 60.05 \\
PURLS~\cite{zhu2024part}         & CVPR 2024 & 79.23 & 40.99 & 71.95 & 52.01 \\
ScoPLe~\cite{zhu2025semantic}    & CVPR 2025 & 84.10 & 52.96 & 74.53 & 52.17 \\
Neuron~\cite{chen2025neuron}     & CVPR 2025 & \textbf{86.90} & \textbf{62.70} & 71.50 & 57.10 \\
TDSM~\cite{do2025bridging}    & ICCV 2025 & 86.49 & 56.03 & 74.15 & \textbf{65.06} \\
\midrule
Ours    & - & 80.04 & 40.17 & 66.01 & 51.85 \\
\bottomrule
\end{tabular}
}
\end{table}

\subsection{Zero-Shot Recognition on NTU RGB+D}
To comprehensively evaluate our model's open-vocabulary capability, we conduct zero-shot action recognition experiments on the widely adopted NTU-60 and NTU-120 datasets. We follow the standard protocol and data splits established by SynSE~\cite{gupta2021syntactically}, such as the 55/5 split on NTU-60, where 55 classes are treated as seen and the remaining 5 as unseen. The results are reported in Table~\ref{tab:zsl_ntu} and compared with a variety of specialized zero-shot action recognition methods. It is important to emphasize a fundamental architectural distinction: most existing zero-shot action recognition approaches adopt a two-stage training paradigm. They first pretrain a skeleton feature extractor, then freeze this encoder and train a separate visual--semantic alignment module. In contrast, our model employs an end-to-end training scheme, jointly optimizing both the feature extractor and the alignment mechanism from scratch using only the seen classes. Despite this more challenging end-to-end formulation, our model achieves competitive performance across all data splits. These results demonstrate that our multi-granularity alignment strategy can effectively learn generalizable representations for open-vocabulary recognition, even without relying on a pretrained, fixed feature extractor.

\subsection{Effect of Pre-Trained Text Encoders}
Table~\ref{tab:text_encoders} investigates the influence of different pre-trained text encoders for generating action label embeddings. Both MPNet, a BERT-based variant, and CLIP demonstrate strong performance. This suggests that the visual encoder can benefit from text embeddings derived from either pure language or vision-language pretraining.

\subsection{Ablation of Alignment Modules (SSA \& FGA)}
To investigate the individual contributions of the Stream-Specific Alignment (SSA) and Fine-Grained Alignment (FGA) modules, we conduct an ablation study by progressively removing them from our full model. The experiments are performed on the J+B+M modality. As shown in Table~\ref{tab:ablation_modules}, both modules provide significant and complementary performance gains. The results clearly demonstrate that both SSA and FGA are crucial for achieving optimal performance. Removing either module leads to a noticeable drop in accuracy across all datasets. Notably, the improvements are particularly pronounced in the more challenging medium-shot and few-shot categories of the HumanML3D benchmark, highlighting the importance of these alignment strategies for learning robust representations that generalize to rare actions.

\begin{table}[t]
  \centering
  \caption{Effect of various pre-trained text encoders on NTU-60. Several widely used language and vision--language encoders, including MPNet and CLIP variants, are evaluated.}
  \label{tab:text_encoders}
  \resizebox{0.75\columnwidth}{!}{
    \begin{tabular}{lcc}
    \toprule
    Text encoder & NTU-60 (3D) & NTU-60 (2D) \\
    \midrule
    MPNet-v1 & 84.41 & 87.39 \\
    MPNet-v2 & 84.61 & 87.40 \\
    CLIP-L/14 & 84.50  & 87.57 \\
    CLIP-L/14@336 & \textbf{84.70} & \textbf{87.58} \\
    \bottomrule
    \end{tabular}}
\end{table}

\begin{table}[t]
\centering
\caption{Ablation study of the SSA and FGA modules on the J+B+M modality. Both components contribute significantly to the final performance.}
\label{tab:ablation_modules}
\resizebox{\columnwidth}{!}{
\begin{tabular}{l|cc|cccc}
\toprule
\multirow{2}{*}{Model Variant} & NTU-60 & NTU-60 & \multicolumn{4}{c}{HumanML3D} \\
& (3D) & (2D) & Overall & Many-shot & Medium-shot & Few-shot \\
\midrule
Base & 85.13 & 88.76 & 55.81 & 68.32 & 45.19 & 36.57 \\
Ours w/o FGA & 86.51 & 89.12 & 60.26 & 71.23 & 50.90 & 47.03 \\
Ours w/o SSA & 86.08 & 89.06 & 60.42 & 71.34 & 51.65 & 48.11 \\
\midrule
Ours & \textbf{87.01} & \textbf{90.10} & \textbf{61.23} & \textbf{71.51} & \textbf{52.28} & \textbf{49.36} \\
\bottomrule
\end{tabular}
}
\end{table}

\end{document}